\documentclass[final]{cvpr}

\usepackage[T1]{fontenc}

\usepackage{times}
\usepackage{epsfig}
\usepackage{graphicx}
\usepackage{amsmath}
\usepackage{amssymb}
\usepackage{enumitem}
\usepackage[numbers,sort]{natbib}

\usepackage{wrapfig,lipsum,booktabs,multirow}
\usepackage[table]{xcolor}
\usepackage{subcaption}
\usepackage[pagebackref=true,breaklinks=true,colorlinks,bookmarks=false]{hyperref}
\newcommand{\notation}[1]{\ensuremath{#1}\xspace}

\newcommand{\Pixel}{\notation{n}}
\newcommand{\Reference}{\notation{Y[\Pixel]}}
\newcommand{\Flash}{\notation{X_{f}[\Pixel]}}
\newcommand{\NoFlash}{\notation{X_{nf}[\Pixel]}}

\newcommand{\kernelA}{\notation{A}}
\newcommand{\kernelB}{\notation{B}}

\newcommand{\myparagraph}[1]{\vspace{0.5em}\noindent\textbf{#1.}\xspace}

\begin{document}

\title{Deep Denoising of Flash and No-Flash Pairs\\for Photography in Low-Light Environments}
\author{Zhihao Xia$^1$, Micha\"el Gharbi$^2$, Federico Perazzi$^3$, Kalyan Sunkavalli$^2$, Ayan Chakrabarti$^1$\\
  $^1$Washington University in St.\ Louis~~~~~~$^2$Adobe Research~~~~~~$^{3}$Facebook\\
  {\tt\small \{zhihao.xia,ayan\}@wustl.edu, \{mgharbi,sunkaval\}@adobe.com, fperazzi@fb.com}
}

\maketitle

\begin{abstract}
  We introduce a neural network-based method to denoise pairs of images taken in
  quick succession, with and without a flash, in low-light environments. Our goal
  is to produce a high-quality rendering of the scene that preserves the color
  and mood from the ambient illumination of the noisy no-flash image, while
  recovering surface texture and detail revealed by the flash. Our network
  outputs a gain map and a field of kernels, the latter obtained by linearly
  mixing elements of a per-image low-rank kernel basis. We first apply the
  kernel field to the no-flash image, and then multiply the result with the gain
  map to create the final output. We show our network effectively learns to
  produce high-quality images by combining a smoothed out estimate of the
  scene's ambient appearance from the no-flash image, with high-frequency albedo
  details extracted from the flash input. Our experiments show significant
  improvements over alternative captures without a flash, and baseline denoisers
  that use flash no-flash pairs. In particular, our method produces images that
  are both noise-free and contain accurate ambient colors without the sharp
  shadows or strong specular highlights visible in the flash image.
\end{abstract}

\section{Introduction}\label{sec:introduction}

Flash photography has long been a convenient way to capture high-quality images
in low-light conditions. A flash illuminates the scene with a bright burst of light at the time of
exposure, allowing the camera to acquire a photograph with a much higher
signal-to-noise ratio than would be possible under the dim ambient lighting
alone and without introducing any motion or defocus blur.
The flash addresses the problem of limited illumination at its
root---by adding light to the scene.
However, flash illumination is not without drawbacks.
An on-camera flash often creates unappealing flat shading and harsh shadows, resulting in images that fail to
capture the true mood and ambience of the scene.

\begin{figure}[!t]
  \centering
  \includegraphics[width=\columnwidth]{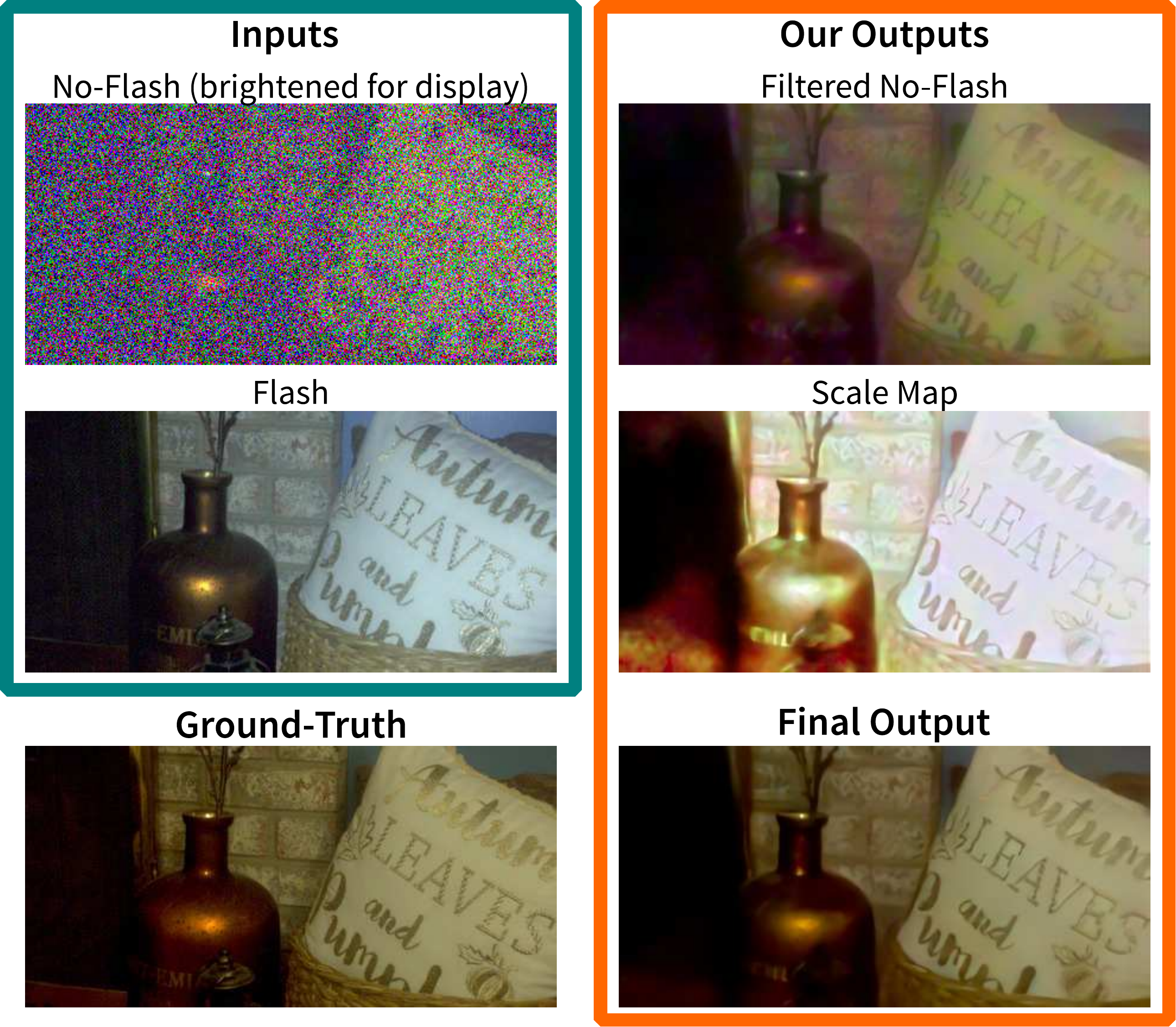}
  \caption{\label{fig:teaser} Given a pair of images of low-light scenes
    captured with and without a flash (left), our method produces a high-quality
    image of the scene under ambient lighting (right). This output is generated
    by filtering the no-flash image with a predicted field of kernels---to
    capture a smoothed stimate of scene appearance under ambient lighting,
    followed by multiplication with a scale map that introduces high-frequency
    detail illuminated by the flash. }
\end{figure}

Researchers have considered combining pairs of flash and no-flash
images---captured in quick succession with and without the flash---to create a
single enhanced photograph that is both noise-free and accurately represents the
scene under ambient lighting. This is achieved by merging information about the
ambient scene appearance from the noisy no-flash image, with high-frequency
surface image details revealed by the flash~\cite{Petschnigg04,EisemannD04}.
However, these methods assume moderate levels of noise in the no-flash image,
and that the flash and no-flash pair are, or can be, aligned.

In extremely low light, the no-flash image can be very noisy, especially 
when using mobile phone cameras with small apertures. This precludes
the use of traditional flash/no-flash methods, since the noise obscures even the
low-frequency shading information in the no-flash image and makes automatic
alignment of the pair unreliable. In comparison, modern neural network-based
denoising methods~\cite{ircnn,zhang2019residual,xia2018identifying,chen2018dark}
can produce reasonable estimates from a noisy no-flash image alone---although at
high noise-levels, they still struggle to reconstruct high-frequency detail.

In this work, we leverage both the ability of modern neural networks to encode
strong natural image priors, and the unique combination of appearance
information available in a flash and no-flash image pair. Specifically, we
consider the task of producing a high-quality image of the scene under ambient
lighting given a flash and no-flash pair as input. We focus on extremely
low-light scenes such that the no-flash image shows significant noise, and the
appearance of the no-flash image is entirely dominated by the flash
illumination. We further assume unknown geometric misalignment between the image
pair, due to camera movement typically observed in hand-held burst
photography~\cite{wronski2019handheld}.

Under these conditions, we train a deep neural network to take noisy, misaligned
no-flash/flash image pairs as input, and output a denoised image of the scene
under the scene's ambient illumination. Rather than directly predicting the
denoised image, our network outputs a kernel field used to filter the no-flash
image, and a scale map that is multiplied with this filtered output to
incorporate high-frequency image details from the flash image. To use the
regularizing effect of kernels to effectively filter out the high levels of noise
in the no-flash input while overcoming its significant memory and computational
costs~\cite{mkpn}, our network combines a recent kernel basis prediction 
approach~\cite{XiaPGSC20} with efficient kernel up-sampling. The use of a scale
map is inspired by classical flash/no-flash approaches~\cite{EisemannD04, Petschnigg04}
that adopt multiplicative combination based on a view of factorizing images into
albedo and shading, where the former is common across the input pair while the latter is not.

We evaluate our approach extensively under different ambient light levels and
spatial misalignment, and demonstrate state-of-the-art results for low-light
denoising (see example result in Fig.~\ref{fig:teaser}). Our method outperforms
denoising without a flash---when using a single or burst of two no-flash images.
This demonstrates that a flash input, despite often representing drastically
different shading, is still informative towards ambient appearance. Our method
also outperforms other standard denoising approaches trained directly on
flash/no-flash pairs, highlighting the importance of the formulation and design
of our network architecture.

Code and pre-trained models for our method are available at \url{https://www.cse.wustl.edu/~zhihao.xia/deepfnf/}.

\section{Related Work}\label{sec:rw}

\noindent\textbf{Image Denoising.}
Early works reduced image noise using regularization schemes like
sparse-coding~\cite{LiL09} and low-rank factorization~\cite{GuoZZL16} to model
the local statistics of natural images. Other classical approaches have
exploited the recurrence of natural image patterns, averaging pixels with
similar local neighborhoods \cite{BuadesCM05, PeronaM90, TomasiM98,
  LindenbaumFB94,Rudin92,Yaroslavsky85,Donoho95}. Current state-of the-art
denoisers use deep neural networks. Burger~\etal~\cite{mlp} were the first to
show the ability of even shallow multi-layer perceptrons to to outperform
traditional methods such as BM3D~\cite{bm3d}, and more recent approaches
utilizing deeper networks and complex architectures~\cite{dncnn, ircnn,
  liu2018non, zhang2019residual,xia2018identifying,xie2012denoising} have since
led to further improvements in reconstruction accuracy.

\myparagraph{Burst Image Denoising} Burst imaging can achieve impressive
denoising results, by capturing multiple frames in quick succession.
Recent burst denoising algorithms have focused on circumventing the frame
misalignment that exists in a real burst. Some methods estimate pixel-wise
displacement~\cite{hdrplus,liu2014fast,heide2016proximal, flexisp,
  kokkinos2019iterative}; others only require coarse global registration, and
rely on neural networks to account for the remaining displacement. Amongst the
latter group, kernel prediction networks~\cite{burstkpn,dburst,mkpn} have
demonstrated superior ability to recover from misalignment. Our method
builds upon Xia~\etal~\cite{XiaPGSC20}, which predicts a low-dimensional kernel
decomposition, using a linear basis and mixing coefficients, to efficiently
realize larger kernels, and to induce regularization leading to improved
reconstructions. However, unlike Xia~\etal~\cite{XiaPGSC20} that operate on a
burst of images, our approach works on misaligned \emph{flash} and
\emph{no-flash} image pairs, and uses a scale map rather than filtering to
extract information from our no-flash input, and an up-sampling approach to
realize even larger kernels.

\begin{figure*}[!t]
  \centering
  \includegraphics[width=\linewidth]{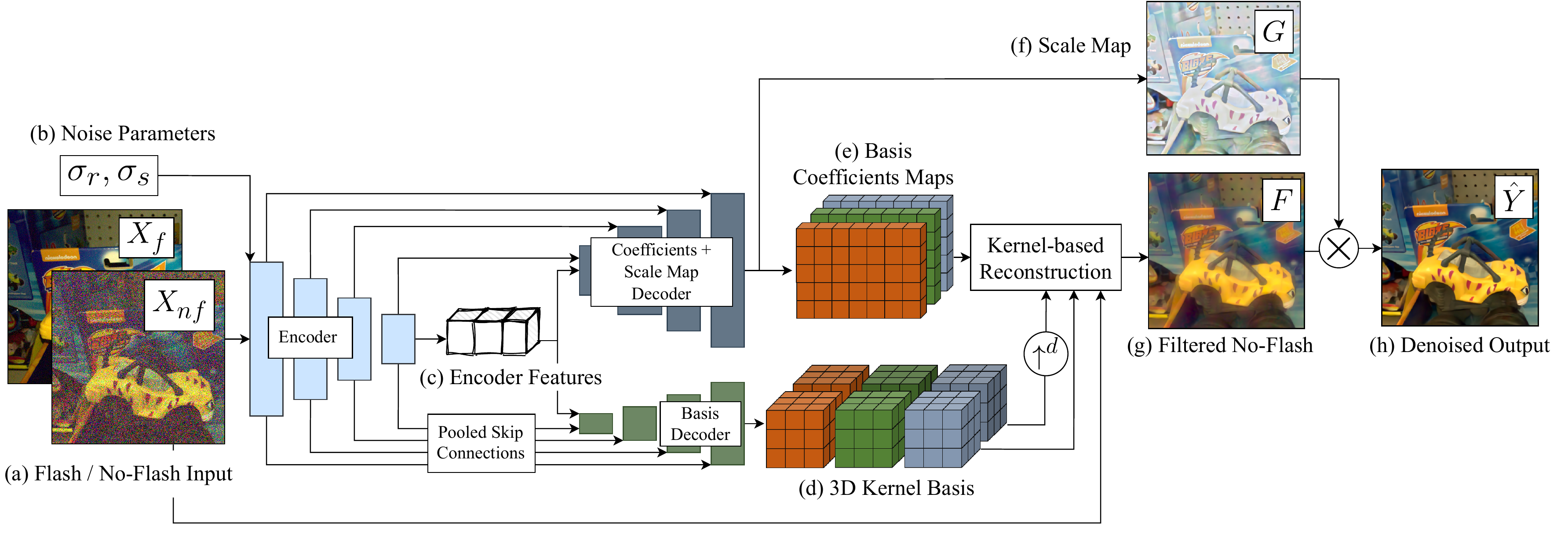}
  \caption{\label{fig:network} {\bf System Overview.} The denoising network
    takes as input a pair of flash, no-flash images (a) together with the noise
    parameters (b). After encoding, the resulting features (c) are decoded into
    a multi-scale basis (d), a set of pixel-wise coefficients (e) and a scale
    map (f). The no-flash image is filtered using the reconstructed kernels (g)
    and multiplied by the scale map to produce the final denoised output (h).}
\end{figure*}

\myparagraph{Flash Denoising} Flash photography enables the capture of low-noise
images in low-light environments using short exposure times and low ISO
settings. But, the additional source flash light drastically changes the mood of
the scene captured. To remedy this while benefiting from the flash image's
higher signal-to-noise ratio, several approaches have used the flash as
reference to denoise a noisy ambient (no flash) image.
Petschnigg~\etal~\cite{Petschnigg04} and Eisemann~\etal~\cite{EisemannD04} use
the flash photo to guide a joint-bilateral filter that denoises the ambient
image, transferring high-frequency content from the flash photo. Krishnan
and Fergus~\cite{KrishnanF09} exploit the correlations between dark
flash images and visible light to denoise the ambient image and restore fine
details. Yan~\etal~\cite{yan2013crossfield} combine gradients from the flash 
image with the no-flash image for denoising. These methods all use hand-crafted
heuristics to decide which image features to preserve from each of the flash and
no-flash inputs.

More recent work~\cite{LiHA016, WangXBC19} have replaced these
heuristics with deep neural networks. Li~\etal~\cite{LiHA016} use the (aligned) 
flash photo as guidance to denoise ambient images. Wang~\etal~\cite{WangXBC19} 
address some of the shortcomings of dark flash photography by adding a stereo RGB 
image to the capture setup. After being registered and aligned the two images are 
fused using recent techniques for hyperspectral image restoration and fast image
enhancement~\cite{ChenXK17,GharbiCBHD17}. Unlike these methods, our approach handles 
motion between the flash/no-flash pairs and large noise levels by using large, 
learned denoising kernels robust to misalignment.

Beyond denoising, flash photography has also been used for other applications
such as deblurring~\cite{ZhuoGS10}, shape estimation~\cite{CaoWSGZM20}, and to
separate shading from different ambient illuminants present in a scene~\cite{HuiSHS18}.

\section{Proposed Approach}\label{sec:method}
Our goal is to estimate a noise-free color image $\Reference\in\mathbb{R}^{3}$
of the scene under ambient illumination, from a pair of flash and no-flash
images $\Flash\in\mathbb{R}^{3}$ and $\NoFlash\in\mathbb{R}^{3}$, where
$\Pixel\in \mathbb{Z}^{2}$ denotes the pixel location. Since both images are of
the same scene, they represent observations of the same surfaces, with the same
material properties, but under different illuminations, and with a potential
change in viewpoint due to hand motion between the two shots.

\subsection{Observation Model and Problem Formulation}\label{sec:formulation}
In a chosen reference frame, we denote the appearance of the scene under
ambient-only illumination as $S_{a}[n]\in\mathbb{R}^{3}$, and under flash-only
illumination as $S_{f}[n]\in\mathbb{R}^{3}$. Further, we model the geometric
transformations from the reference to the flash and no-flash images as 2D warps
$\mathcal{T}_{f}(n)$ and $\mathcal{T}_{nf}(n)$, respectively. Then, the
\emph{noise-free} versions $\tilde{X}_{nf}[n]$ and $\tilde{X}_{f}[n]$ of our
no-flash and flash inputs are given by:
\begin{align}
  \label{eq:nonzdef}
  \tilde{X}_{nf}[\mathcal{T}_{nf}(n)] &= \ S_{a}[n], \notag\\
  \tilde{X}_{f}[\mathcal{T}_{f}(n)] &= \alpha_{f}\ \ (S_{f}[n] + S_{a}[n]),
\end{align}
where $\alpha_{f} \leq 1$ is a scalar that captures the effect of a possibly
shorter exposure time for the flash image. Note that since the flash is
typically much brighter than the ambient lighting ($S_{f}[n] \gg S_{a}[n]$), the
contribution of the flash-only appearance is dominant in the flash image
$\tilde{X}_{f}[n]$.

As in~\cite{burstkpn,XiaPGSC20}, we assume a heteroscedastic Gaussian noise
model~\cite{foi2008practical} to account for both read and shot sensor noise.
The observed input flash and no-flash pair relate to their ideal noise-free
version (Eq.~\eqref{eq:nonzdef}) as:
\begin{align}
  \label{eq:nzdef}
   X_{f}[n] &\sim \mathcal{N}(\tilde{X}_{f}[n], \sigma_{r}^{2} + \sigma_{s}^{2}\tilde{X}_{f}[n]),\notag\\
  X_{nf}[n] &\sim \mathcal{N}(\tilde{X}_{nf}[n], \sigma_{r}^{2} + \sigma_{s}^{2}\tilde{X}_{nf}[n]),
\end{align}
where $\sigma_{r}^{2}, \sigma_{s}^{2}$ are read and shot noise parameters, which
we assume are known. Given $X_{f}[n]$ and $X_{nf}[n],$ and the values of
$\sigma_{r}^{2}$ and $\sigma_{s}^{2}$, we seek to estimate $Y[n] := S_{a}[n]$.

\myparagraph{Flash vs.\ No-flash as reference} Note that in formulation above,
we make a distinction between the target output $Y[n]$ and the noise-free
no-flash image $\tilde{X}_{nf}[n]$, because they differ by the warp $T_{nf}(n)$.
We may wish to use either of the two inputs (flash or no-flash) as the
\emph{geometric reference}. If for instance, the no-flash image is the
reference, we assume $T_{nf}(n)=n$ is the identity transformation, and
$Y[n]=\tilde{X}_{{nf}}[n]$. Conversely, if we choose the flash image as
reference, $T_{f}(n)=n$ is choosen to be the identity mapping. In
Section~\ref{sec:exp}, we analyze the effect of this design choice on the output
image quality, finding that in most settings, the choice of the no-flash image
as reference yields more accurate reconstructions on average.

\subsection{Enhancement Network}\label{sec:network}
We use the basis prediction approach of Xia et al.~\cite{XiaPGSC20}, which was
designed for burst denoising, as the starting point for our model design. Our
network differs in two crucial aspects: (a) rather than predicting kernels to
filter both the flash and no-flash inputs and summing the result, we filter only
the no-flash image and multiply a predicted per-pixel three-channel scale map to
form our final output; and (b) we propose an efficient approach to predict
larger kernels through upsampling, which is necessary in our setting, because we
are filtering a single, highly noisy image. We show an overview of our approach
in Figure~\ref{fig:network}, and include a complete description of our network
in the supplement.

\subsubsection{Input data}
Our network takes a twelve channel tensor as input, with six channels containing
the observed flash ${X}_{f}$ and no-flash ${X}_{nf}$ pair (Equation~\ref{eq:nzdef}) themselves, and another six encoding the
expected per-pixel standard deviation of noise in these inputs, computed using the (known) values of
$\sigma_{s}^{2}$ and $\sigma_{r}^{2}$ and the observed noisy intensities as:
$\sqrt{\sigma_{r}^{2} + \sigma_{s}^{2}\max(0,X^{i}[n])}$, for each channel
$i\in\{R,G,B\}$ and $X=X_{f}$ and $X_{nf}$.

\subsubsection{Predicting a global kernel basis}
Like~\cite{XiaPGSC20}, our network features a common encoder whose output is fed
to two decoders. The first decoder outputs a global low-rank kernel basis.
Unlike~\cite{XiaPGSC20}, we do not constrain our kernels to be positive and
unit-normalized. The second decoder outputs per-pixel mixing coefficients to
combine the predicted basis elements and form per-pixel kernels. Another
departure from~\cite{XiaPGSC20}, the second decode also outputs a 3-channel
scale map. We include skip-connections from the encoder to both decoders, using
global pooling for connections to the basis decoder as in~\cite{XiaPGSC20}.

\subsubsection{Large kernels by interpolation}
A key innovation in our method over~\cite{XiaPGSC20} is that our basis encodes
larger kernels using a 2-scale representation and an interpolation-based
reconstruction scheme. This is crucial in our application where these kernels
are used to smooth only one image---the noisy no-flash input---rather than a
burst of images as in~\cite{XiaPGSC20}.

Specifically, our basis decoder outputs a set of $J$ basis elements, each
consisting of a pair of three-channel kernels
$\{(\kernelA_{j}, \kernelB_{j})\}_{j=1}^{J}$, where each
$\kernelA_{j},\kernelB_{j} \in \mathbb{R}^{K\times K\times 3}$. We interpret the
second kernel $\kernelB_{j}$ of each pair as a low-frequency term: a large
kernel downsampled by a factor $d$, with an effective
$((K-1)*d+1)\times((K-1)*d+1)$ footprint. The $j^{th}$ element of our basis is
then given by $\kernelA_{j} + (\kernelB_{j}\uparrow^{d})$, where $\uparrow^{d}$
denotes bilinear upsampling by a factor $d$. So that $\kernelA_{j}$ can add fine
high-frequency details to the kernel center. In our experiments, we use a basis
with $J=90$ kernels, with a base size $K=15$ and upsampling factor of $d=4$
resulting in an effective kernel size of $57\times 57$.

\subsubsection{Final reconstruction}
Denoting the per-pixel coefficients from the second decoder as
$\{c_{j}[n]\}_{j=1}^{J}$, we first filter the no-flash input image as:
\begin{equation}
  \label{eq:filt1}
  F[n] = \sum_{j=1}^{J} c_{j}[n] \left(X_{nf} * (\kernelA_{j} + \kernelB_{j}\uparrow^{d})\right)[n].
\end{equation}
where $*$ denotes per-channel convolution between three-channel images and
kernels. Note that the filtering with upsampled kernels can be carried out
efficiently, by pre-filtering the no-flash image and using dilated convolutions:
\begin{equation}
  \label{eq:filt2}
  F[n] = \sum_{j=1}^{J} c_{j}[n] \left( (X_{nf}*\kernelA_{j})[n] + (X^{h}_{nf} *_{d}\kernelB_{j})[n]\right),
\end{equation}
where $X^{h}_{nf}[n] = (X_{nf}*h)[n]$ is the result of smoothing the no-flash
input with a $(2d-1)\times(2d-1)$ tent kernel $h[n]$, and $*_{d}$ represents
dilated convolution with a factor of $d$.

\myparagraph{Recovering high-frequency detail with a scale map} The result
$F[n]$ of this filtering step will typically encode a noise-free (and in the
case of the flash as reference, an aligned) estimate of scene appearance under
ambient illumination. However, due to the lower signal-to-noise ratio of
$X_{nf}$, this filtering step cannot recover the high-frequency details that are
illuminated only resolved in the flash image. To recover these, our full-pixel
decoder also produces a scale map $G[n] \in \mathbb{R}^{3}$. Our final output
$\hat{Y}[n]$ is given by the element-wise product of this scale map and the
filtered no-flash image:
\begin{equation}
  \label{eq:scale}
  \hat{Y}[n] = F[n]\odot G[n].
\end{equation}
This formulation is inspired by classic flash/no-flashing denoising
methods~\cite{Petschnigg04,EisemannD04} that add high-frequency details from the
flash image in the \emph{log domain}, i.e., corresponding to a product in our
linear domain. In Section~\ref{sec:exp}, we show this outperforms the
alternative of using kernels to jointly denoise the no-flash and flash images.

\subsection{Training details}\label{sec:tdetail}

While our network accepts raw linear sensor measurements as input and produces
an estimate of linear intensities in $Y[n]$, it is trained to maximize image
quality in a color and gamma-corrected sRGB space. In particular, we assume that
for each training sample $(X_{f}^{t}, X_{nf}^{t}, Y^{t})$, we also have a scalar
gain $\alpha^{t}$ (representing a desired target brightness level), and a
$3\times 3$ color transform matrix $C^{t}$ based on camera sensor parameters and
white-balance settings, such that the mapping to sRGB is given by
$f_{t}(Y[n]) = \gamma(\alpha C^{t}\hat{Y}[n])$, where $\gamma(\cdot)$ is a gamma
correction curve.

We train our model to minimize the sum of a squared $L_2$ pixel loss, and a
$L_1$ gradient loss between the estimated and ideal rendered images:
\begin{align}
  \label{eq:lossdef}
  L &= \frac{1}{T} \sum_{t=1}^{T} \|f_{t}(\hat{Y}_{t}) - f_{t}(Y_{t})\|^{2} + \eta |\partial_{x} * (f_{t}(\hat{Y}_{t}) - f_{t}(Y_{t}))| \notag\\&\qquad\qquad\qquad+ \eta|\partial_{y} * (f_{t}(\hat{Y}_{t}) - f_{t}(Y_{t}))|,
\end{align}
where $\partial_{x}$ and $\partial_{y}$ are horizontal and vertical gradient
filters.

We train our model using the Adam optimizer~\cite{kingma2014adam}, beginning
with a learning rate of $10^{-4}$, and going through two learning rate drops
every time validation loss saturates, for a total of roughly 1.5 million
iterations.

\section{Experiments}\label{sec:exp}
We now describe experiments evaluating our approach, and comparing it to
baseline methods for both denoising without a flash input, and to applying
existing network architectures to a flash and no-flash pair. We also include
ablations describing the effect of our kernel interpolation approach, and of
choosing the no-flash vs.\ flash image as geometric reference.

\begin{table*}[!t]
  \centering
  \begin{tabular}{cl c cc c cc c cc c cc}
    \toprule
    &\multirow{2}{*}{\textbf{Method}} && \multicolumn{2}{c}{\bf 100x Dimmed} && \multicolumn{2}{c}{\bf 50x Dimmed} && \multicolumn{2}{c}{\bf 25x Dimmed} && \multicolumn{2}{c}{\bf 12.5x Dimmed}\\
    &&& PSNR & SSIM && PSNR & SSIM && PSNR & SSIM && PSNR & SSIM\\\midrule

    \rowcolor{lightgray}\multicolumn{14}{l}{No-flash input only}\\
    &Our Architecture    && 24.91 dB & 0.779 && 27.23 dB & 0.825 && 29.31 dB & 0.865 && 30.98 dB & 0.895\vspace{0.5em}\\
    \rowcolor{lightgray}\multicolumn{14}{l}{2-frame burst input (no flash)}\\
    &BPN~\cite{XiaPGSC20} && 25.58 dB & 0.796 && 27.75 dB & 0.839 && 29.65 dB & 0.874 && 31.21 dB & 0.899\vspace{0.5em}\\
    \rowcolor{lightgray}\multicolumn{14}{l}{Flash and no-flash input pair}\\
    &Direct Prediction    && 24.80 dB & 0.773 && 27.06 dB & 0.818 && 29.12 dB & 0.857 && 30.84 dB & 0.888\\
    &KPN~\cite{burstkpn}  && 25.87 dB & 0.815 && 27.94 dB & 0.852 && 29.69 dB & 0.880 && 31.21 dB & 0.901\\
    &BPN~\cite{XiaPGSC20} && 26.11 dB & 0.815 && 28.04 dB & 0.850 && 29.75 dB & 0.880 && 31.21 dB & 0.901\\
    &\textbf{Ours}        && \bf 26.75 dB & \bf 0.829 && \bf 28.56 dB & \bf 0.860 && \bf 30.14 dB & \bf 0.884 && \bf 31.52 dB & \bf 0.903\\\bottomrule

  \end{tabular}
  \caption{\label{tab:main} {\bf Quantitative results.} Thanks to the richer
    signal provided by the flash input, our method outperforms our single image
    denoising baseline, and a 2-frame burst denoising baseline. Comparisons to
    standard burst denoising approaches adapted to use flash--no-flash pairs
    show that our model architecture with its filtering/scale decomposition and
    larger kernels outperforms previous work. These results hold over a wide
    range of ambient light levels, shown here as dimming factors between the
    low-light no-flash input and a well-lit ground-truth target.}
\end{table*}

\subsection{Preliminaries}\label{sec:data}

\noindent\textbf{Dataset.} We use the dataset of Aksoy~\etal~\cite{AksoyKKPEPM18}, 
which contains 16-bit well-exposed ambient-only and flash-only image pairs. 
These images were crowdsourced from users who were asked to capture images 
with hand-held mobile phones in real-world settings, and roughly had a 0.5-1 
second delay between captures of the pair. We split the dataset as follows: 
2519 images for the training set, 128 for validation and 128 for testing, 
considering $440\times 440$ crops (random crops for training, and fixed central 
crops for validation and testing). We simulate a real low-light capture by dimming 
the linear ambient-only image by dividing with a random factor in $[2, 50]$, 
sampled uniformly in the log domain. This forms our \emph{no-flash} input. 
We increase the exposure of the flash-only image by a constant factor 2, and 
add it to the no-flash input to obtain our \emph{flash} input.

\myparagraph{Misalignment and simulated noise} The image pairs provided by~\cite{AksoyKKPEPM18} 
were automatically aligned by finding correspondences with feature matching. 
Since this would be unrealistic in low-light images, we undo such alignment 
by warping the no-flash or flash image (the other is the reference) 
with a random homography. To obtain the homography parameters, we assume the 
camera's FOV is 90 degrees to get its intrinsic matrix. We perturb 
it with a random 3D rotation uniformly sampled in the range $[-0.5, 0.5]$ degrees 
in each axis, followed by random 2D scaling by a factor uniformly sampled 
in $[0.98, 1.02]$, and a random 2D translation of $[0,2]$ pixels. 
The overall average per-pixel displacement between our flash and no-flash inputs 
ranges up to 20 pixels (Manhattan distance). Note that real-world non-idealities 
like parallax, occlusions, blur, etc. originally present in the data are 
preserved by undoing~\cite{AksoyKKPEPM18}'s alignment.

We use the same noise parameters for the flash and no-flash image, i.e., we
assume they were captured with the same ISO setting. During training, we
randomly sample the noise parameters $\sigma_{r}$ and $\sigma_{s}$ uniformly in
the log-domain in the ranges: $\log(\sigma_r) \in [-3, -2]$ and
$\log(\sigma_s) \in [-4, -2.6]$.

\myparagraph{Losses and metrics} The preprocessing pipeline is executed on the
original linear color space of the camera. To compute losses, we set the desired
gain $\alpha^{t}$ in Sec.~\ref{sec:tdetail} to be the inverse of the factor we
used to dim the image above, since the original images in~\cite{AksoyKKPEPM18}
were well-lit. The database also includes a color transform matrix for each
image which we use as $C^{t}$. We evaluate performance by computing PSNR and
SSIM between the rendered versions of our estimate and the ground-truth.

\myparagraph{Baselines} We compare to denoising without a flash input: using a
single no-flash image denoised by a version of our architecture (without a scale
map), and a burst of two (misaligned) no-flash inputs denoised using the
state-of-the-art burst denoising method of \cite{XiaPGSC20} (which we refer to
as BPN). For flash and no-flash image inputs, we compare our method to other
standard architectures: a direct prediction network which simply regresses to
the denoised output, and burst denoising methods KPN~\cite{burstkpn} and
BPN~\cite{XiaPGSC20} applied to the flash and no-flash pair. All of these
methods were trained on our dataset, and provided information about noise
standard deviation in an identical manner to our method. Further details are
included in the supplement.

\subsection{Evaluation}\label{sec:eval}

We begin by evaluating our method, choosing the ambient image as geometric
reference (i.e., we assume $T_{nf}(n) = n$), on our test set of 128 images, and
comparing it to the various baselines described above. We
fix the noise level to $\log(\sigma_r) = -2.6$ and $\log(\sigma_s) = -3.6$,
sample a random homography for each pair to be applied to the flash image (for
the no-flash burst, this homography is applied to the second no-flash image),
and repeat our evaluation with a discrete set of dimming factors:
$[100, 50, 25, 12.5]$. Note that the factor $100$ lies outside our training
range, and demonstrates the robustness of our method.

\begin{figure*}[p]
  \centering
  \includegraphics[width=0.84\textwidth]{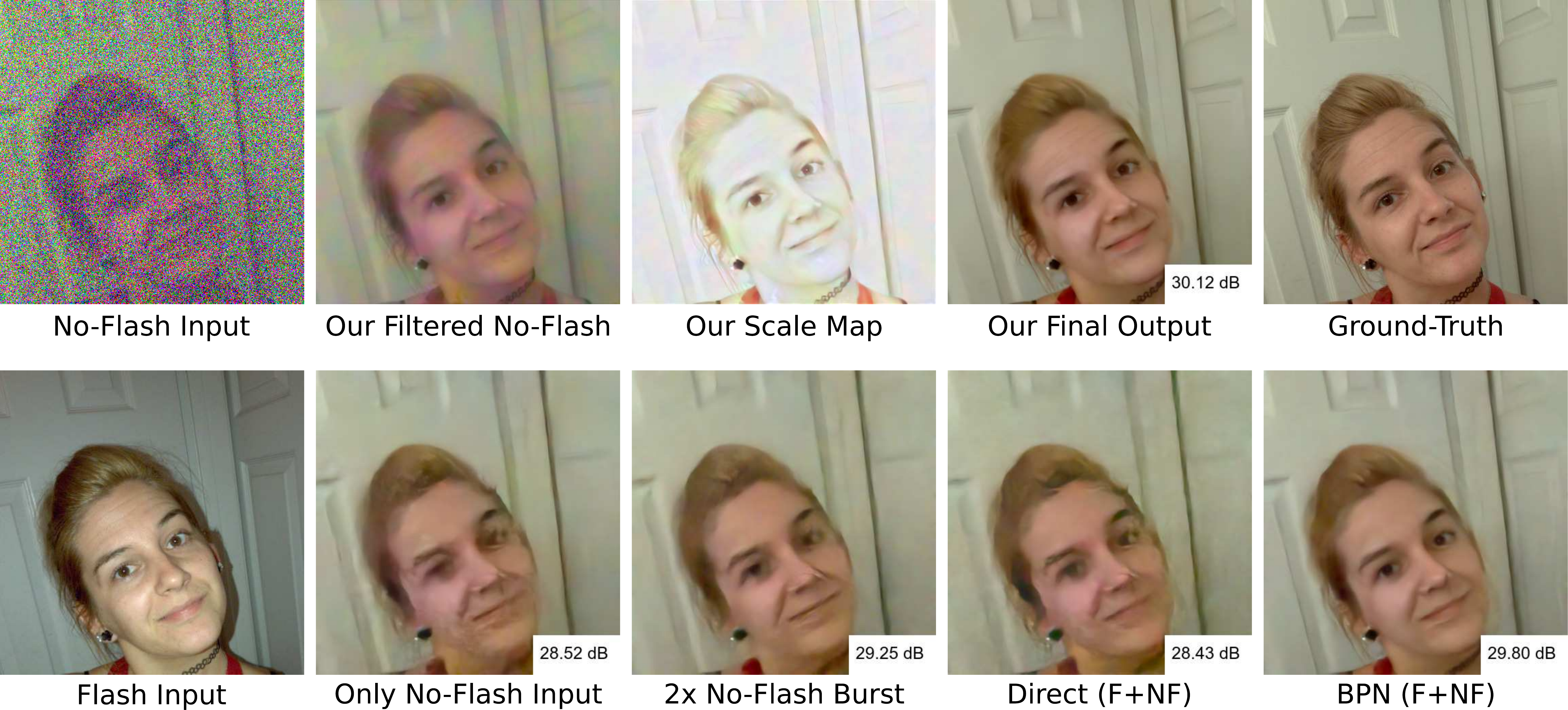}\\~\\
  \includegraphics[width=0.84\textwidth]{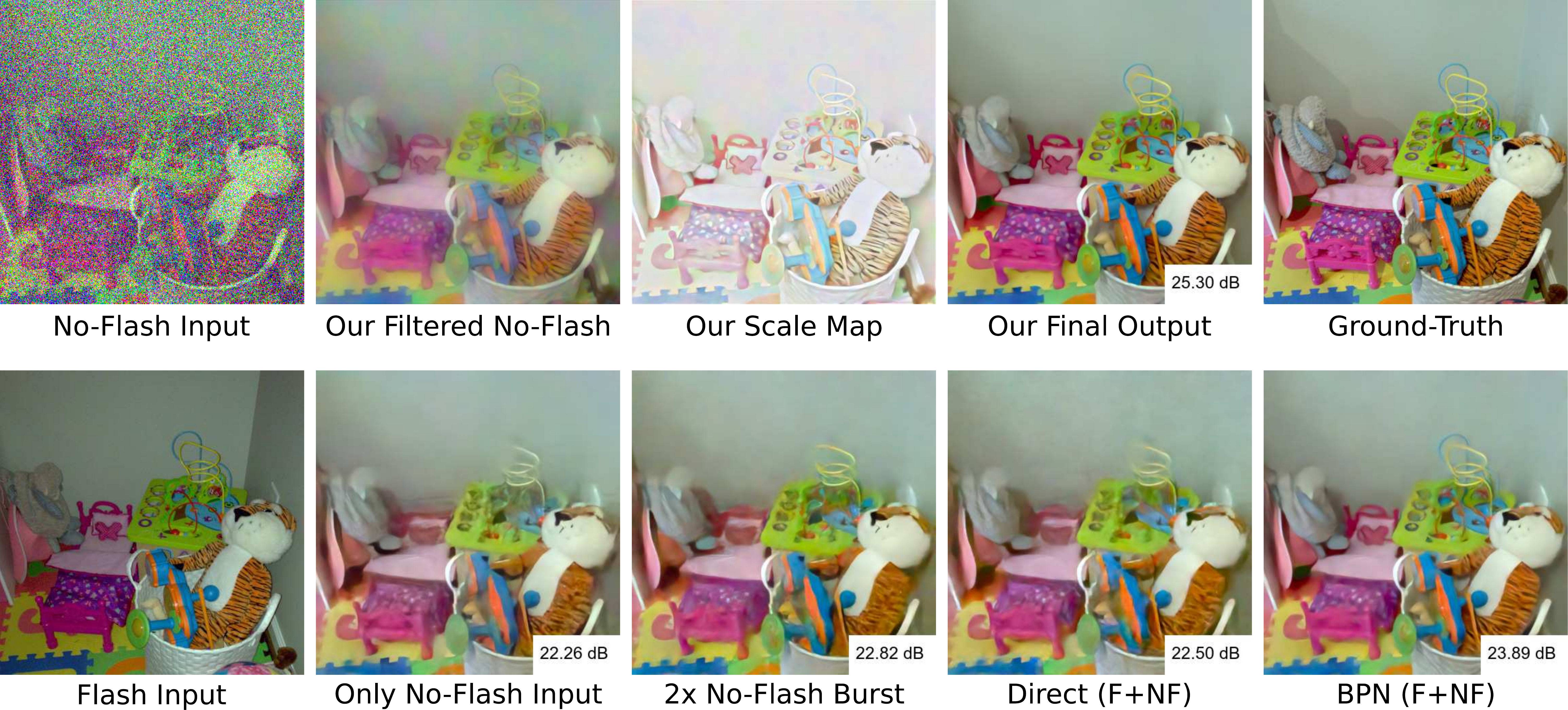}\\~\\
  \includegraphics[width=0.84\textwidth]{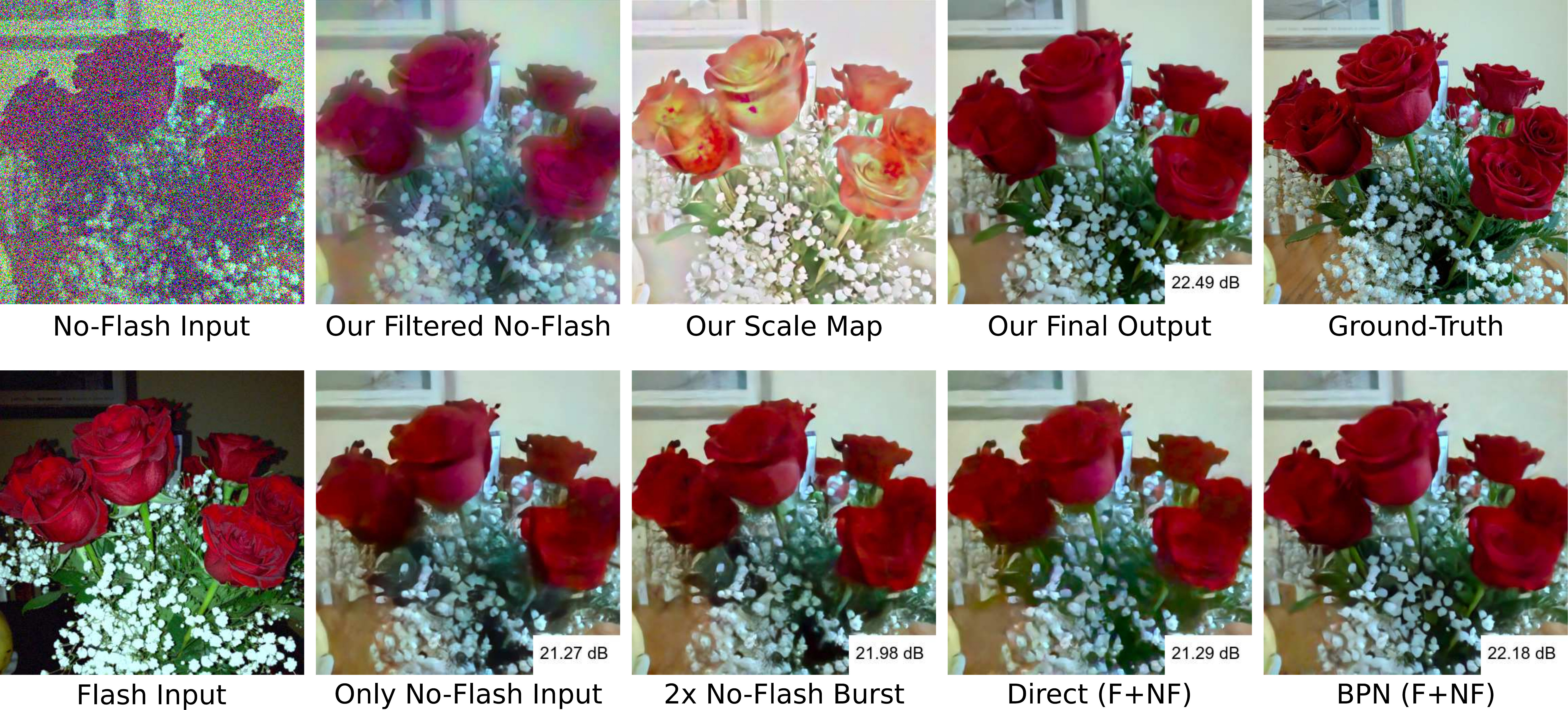}
  \caption{\label{fig:main} {\bf Qualitative comparison.} Our method uses
    flash/no-flash image pairs to denoise low-light images. It produces cleaner
    outputs than baseline flash/no-flash denoisers (\emph{Direct (F+NF)},
    \emph{BPN (F+NF)}), as well as single-image (\emph{Only No-Flash Input}) and
    burst denoisers (\emph{$2\times$ No-Flash Burst}). We also visualize our
    intermediate filtered no-flash image and scale map.}
\end{figure*}

Our method consistently outperforms all methods, regardless of the dimming
factor, as seen by the quantitative results in Table~\ref{tab:main}. We also
include example reconstructions in Fig.~\ref{fig:main}, where we see that our
method reconstructs fine surface detail with higher fidelity than the other
methods.

\begin{figure}[!tb]
  \centering
  \includegraphics[width=.85\columnwidth]{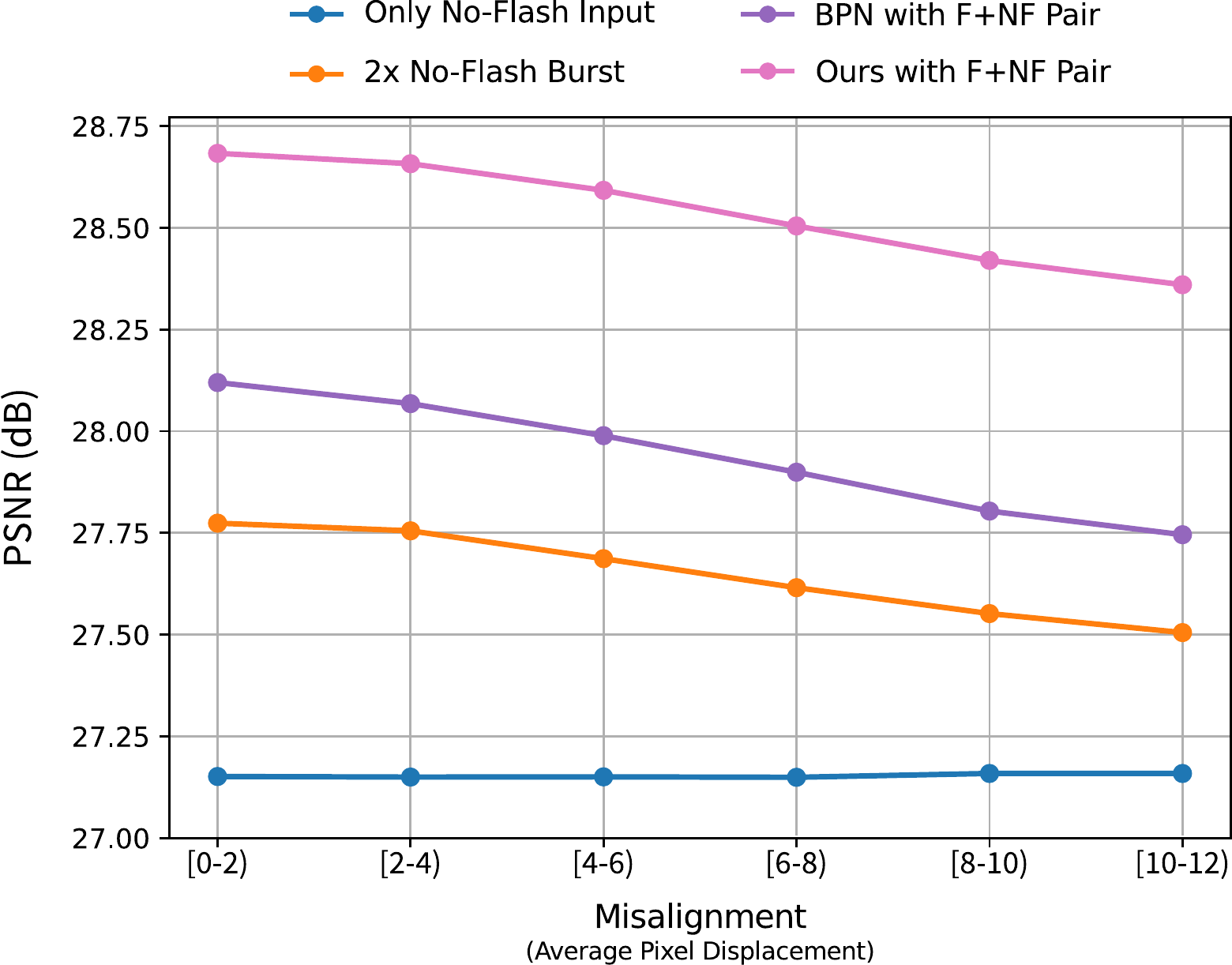}
  \caption{\label{fig:align} {\bf Performance vs.\ misalignment}. We show the
    performance profile of our method and select baselines as a function of
    average displacement between the two frames. Our model consistently delivers
    superior performance and is robust to large misalignment between its inputs.
  }
\end{figure}

In Fig.~\ref{fig:align}, we take a closer look at the effect of misalignment. We
take a subset of 64 flash and no-flash pairs from our test set (all dimmed with
a factor of 50), and evaluate each set with different homographies that cause
different average pixel displacements. We plot the PSNR of reconstruction by
various methods for different degrees of displacement (for the single no-flash
input baseline, these numbers are the same for all displacements). As expected,
the accuracy of all methods decreases with greater misalignment. Nevertheless,
we find that our method consistently outperforms all baselines, including the
single no-flash input even with misalignment greater than 10 pixels.

Additional results, including results on different noise levels and combining 
flash denoising with burst denoising, can be found in the supplemental material.

\subsection{Ablation}

\begin{table}[!tb]
  \centering
  \begin{tabular}{rccc}
    \toprule
    \multirow{2}{*}{Setting} & \bf Flash & \bf No-Flash & \bf No-Flash\\
    &\bf Reference&\bf Reference & \bf w/o $\{B_{j}\}$\\\midrule
    100x dimmed  & \bf 26.83 dB & 26.75 dB & 26.45 dB\\
    50x dimmed   & 28.39 dB & \bf 28.56 dB & 28.42 dB\\
    25x dimmed   & 29.55 dB & \bf 30.14 dB & 30.09 dB\\
    12.5x dimmed & 30.45 dB & \bf 31.52 dB & 31.51 dB\\\bottomrule
  \end{tabular}
  \caption{\label{tab:ablation} {\bf Ablation study.} We compare the performance
    of our method to two ablations. One uses the flash image instead of the
    no-flash image as reference for the geometric transformation. The other uses
    a kernel basis without interpolation, leading to an effective kernel size of
    only $15 \times 15$.}
\end{table}

\begin{figure*}[tbp]
  \centering
  \includegraphics[width=0.76\textwidth]{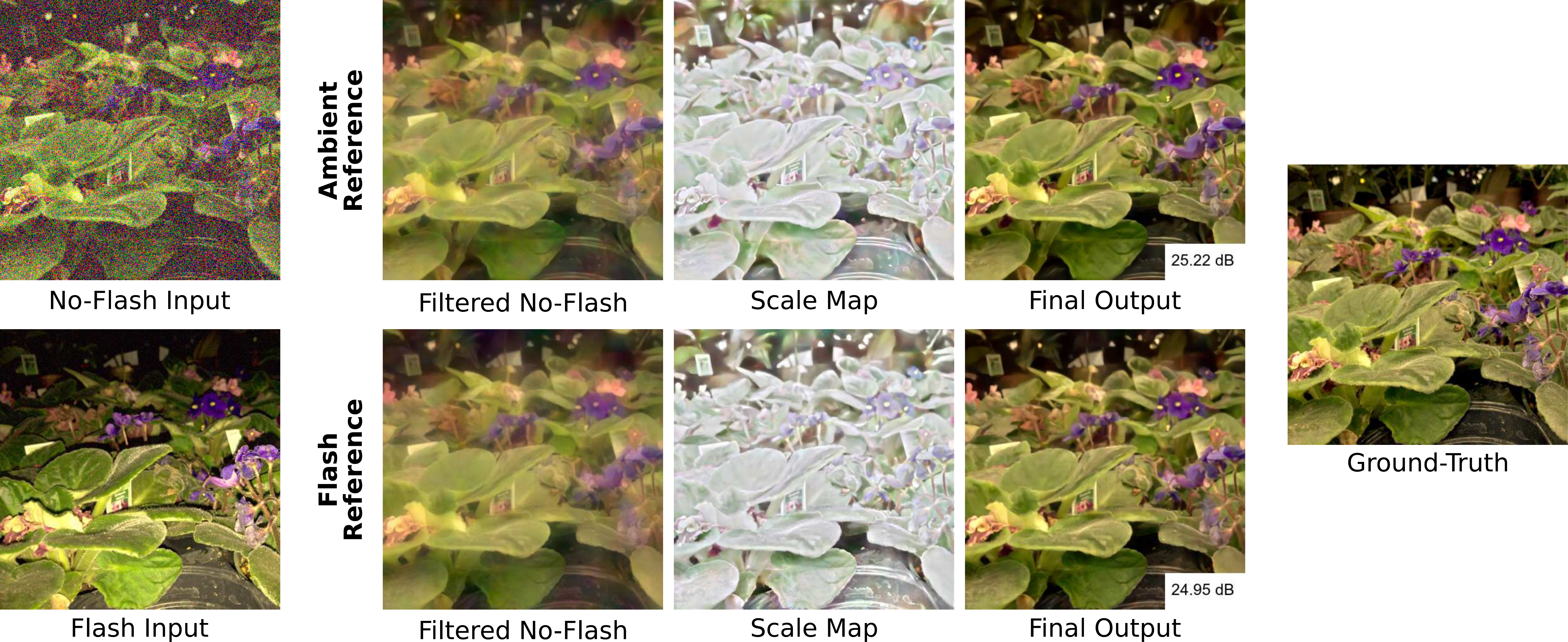}
  \caption{\label{fig:target} {\bf Flash vs.\ no-flash as reference frame.} We
    use the ambient-only image as the reference frame for our reconstruction
    (\emph{top}), i.e.\ the ground truth is aligned to the no-flash image. We
    found this choice leads to a lower error on average, compared to the
    alternative, using the flash as reference (\emph{bottom}). }
\end{figure*}

\begin{figure*}[tbp]
  \centering
  \includegraphics[width=0.76\textwidth]{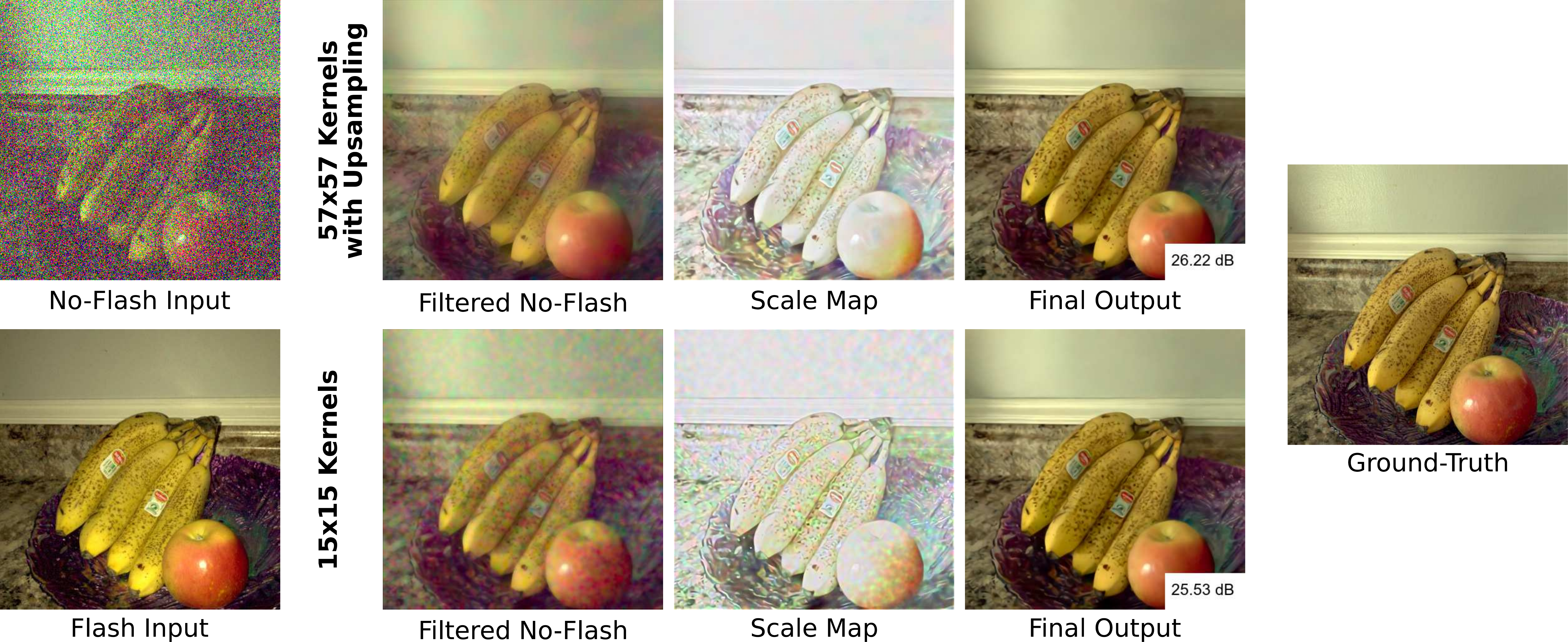}
  \caption{\label{fig:mscale} {\bf Benefit of large kernels.} By using a 2-scale
    kernel decomposition, where the low-pass component is bilinearly upsampled,
    our model (\emph{top}) can better denoise the ambient-only image. This leads
    to reduced residual chroma noise, which makes the scale map more effective
    at recovering fine details. Without it (\emph{bottom}), the kernels are too
    small to effectively denoise the ambient image, so the scale map needs to
    compensate for the residual mid-frequency noise.}
\end{figure*}

Section~\ref{sec:formulation} considered two options for the alignment
reference: with the output geometrically aligned with the flash input, or the
no-flash image. In the previous section, we reported results with the no-flash
input as the reference (for our method, as well as the other methods evaluated
on the flash and no-flash pair). This was based on an evaluation of both
alternatives, which we report in Table.~\ref{tab:ablation}.

We found that except for the lowest light level, the using the no-flash image as
reference yields results that are quantitatively better (this is also true for
the other baselines). However, looking at the actual reconstructions in
Fig.~\ref{fig:target}, we find both images to be of similar visual
quality---with the lower quantitative performance of the flash reference being
largely due to slight, and largely imperceptible, alignment errors in
low-frequency shading. However, as shown in expanded analysis in the supplement,
using the flash image as reference sometimes yields visually sharper results

Table~\ref{tab:ablation} also evaluates the benefit of using larger filters
though our interpolation-based approach. We find that by allowing filters with a
larger footprint ($57\times 57$), our two-scale kernel basis improves denoising
quality, especially at low light levels. As show in Figure~\ref{fig:mscale},
large kernels yield a smoother filtering of the noisy no-flash image, so that
the flash-driven scale map does not need to overcompensate for residual
mid-frequency color noise, leading to better reconstructions in the final
output.

\section{Conclusion}\label{sec:conc}

This paper introduced a method to effectively leverage the unique mix of visual 
information available in a flash and no-flash image pair, and produce high-quality 
images in low-light environments. Our method preserves the warmth and colors of 
the ambient lighting while bringing out fine details thanks to the flash image. 
Drawing on traditional flash/no-flash techniques, our network architecture 
assembles its output from a filtered ambient-only image, and a scale map that 
encoded high-frequency details from the flash. Although it was not trained with 
any intermediate supervision, we found our network automatically learns to carry 
out both the necessary geometric alignment between the frames, and the photometric 
transfer needed to produce state-of-the-art reconstructions. Still, there remain 
situations where flash photography may be too obtrusive. Exploring how our model 
would fare with dark flash imaging~\cite{KrishnanF09,WangXBC19} is an interesting 
avenue for future research.

\myparagraph{Acknowledgments} ZX and AC acknowledge support from NSF award
IIS-1820693, and a gift from Adobe Research.

{\small
\bibliographystyle{ieee_fullname}
\bibliography{refs}

\begin{thebibliography}{10}\itemsep=-1pt

\bibitem{AksoyKKPEPM18}
Yagiz Aksoy, Changil Kim, Petr Kellnhofer, Sylvain Paris, Mohamed~A. Elgharib,
  Marc Pollefeys, and Wojciech Matusik.
\newblock A dataset of flash and ambient illumination pairs from the crowd.
\newblock In {\em Proc.~ECCV}, 2018.

\bibitem{BuadesCM05}
Antoni Buades, Bartomeu Coll, and Jean{-}Michel Morel.
\newblock A non-local algorithm for image denoising.
\newblock In {\em Proc.~CVPR}, 2005.

\bibitem{mlp}
Harold~C Burger, Christian~J Schuler, and Stefan Harmeling.
\newblock Image denoising: Can plain neural networks compete with bm3d?
\newblock In {\em Proc.~CVPR}, 2012.

\bibitem{CaoWSGZM20}
Xu Cao, Michael Waechter, Boxin Shi, Ye Gao, Bo Zheng, and Yasuyuki Matsushita.
\newblock Stereoscopic flash and no-flash photography for shape and albedo
  recovery.
\newblock In {\em Proc.~CVPR}, 2020.

\bibitem{chen2018dark}
Chen Chen, Qifeng Chen, Jia Xu, and Vladlen Koltun.
\newblock Learning to see in the dark.
\newblock In {\em Proc.~CVPR}, 2018.

\bibitem{ChenXK17}
Qifeng Chen, Jia Xu, and Vladlen Koltun.
\newblock Fast image processing with fully-convolutional networks.
\newblock In {\em Proc.~ICCV}, 2017.

\bibitem{bm3d}
Kostadin Dabov, Alessandro Foi, Vladimir Katkovnik, and Karen Egiazarian.
\newblock Color image denoising via sparse 3d collaborative filtering with
  grouping constraint in luminance-chrominance space.
\newblock In {\em Proc.~ICIP}, 2007.

\bibitem{Donoho95}
David~L. Donoho.
\newblock De-noising by soft-thresholding.
\newblock {\em {IEEE} Transactions on Information Theory}, 41(3):613--627,
  1995.

\bibitem{EisemannD04}
Elmar Eisemann and Fr{\'{e}}do Durand.
\newblock Flash photography enhancement via intrinsic relighting.
\newblock {\em ACM Transactions on Graphics (TOG)}, 23(3):673--678, 2004.

\bibitem{foi2008practical}
Alessandro Foi, Mejdi Trimeche, Vladimir Katkovnik, and Karen Egiazarian.
\newblock Practical poissonian-gaussian noise modeling and fitting for
  single-image raw-data.
\newblock {\em IEEE Transactions on Image Processing}, 17(10):1737--1754, 2008.

\bibitem{GharbiCBHD17}
Micha{\"{e}}l Gharbi, Jiawen Chen, Jonathan~T. Barron, Samuel~W. Hasinoff, and
  Fr{\'{e}}do Durand.
\newblock Deep bilateral learning for real-time image enhancement.
\newblock {\em ACM Transactions on Graphics (TOG)}, 36(4):118:1--118:12, 2017.

\bibitem{dburst}
Cl{\'e}ment Godard, Kevin Matzen, and Matt Uyttendaele.
\newblock Deep burst denoising.
\newblock In {\em Proc.~ECCV}, 2018.

\bibitem{GuoZZL16}
Qiang Guo, Caiming Zhang, Yunfeng Zhang, and Hui Liu.
\newblock An efficient svd-based method for image denoising.
\newblock {\em {IEEE} Transactions on Circuits and Systems for Video
  Technology}, 26(5):868--880, 2016.

\bibitem{hdrplus}
Samuel~W Hasinoff, Dillon Sharlet, Ryan Geiss, Andrew Adams, Jonathan~T Barron,
  Florian Kainz, Jiawen Chen, and Marc Levoy.
\newblock Burst photography for high dynamic range and low-light imaging on
  mobile cameras.
\newblock {\em ACM Transactions on Graphics (TOG)}, 35(6):192, 2016.

\bibitem{heide2016proximal}
Felix Heide, Steven Diamond, Matthias Nie{\ss}ner, Jonathan Ragan-Kelley,
  Wolfgang Heidrich, and Gordon Wetzstein.
\newblock Proximal: Efficient image optimization using proximal algorithms.
\newblock {\em ACM Transactions on Graphics (TOG)}, 35(4):84, 2016.

\bibitem{flexisp}
Felix Heide, Markus Steinberger, Yun-Ta Tsai, Mushfiqur Rouf, Dawid
  Paj{\k{a}}k, Dikpal Reddy, Orazio Gallo, Jing Liu, Wolfgang Heidrich, Karen
  Egiazarian, et~al.
\newblock Flexisp: A flexible camera image processing framework.
\newblock {\em ACM Transactions on Graphics (TOG)}, 33(6):231, 2014.

\bibitem{HuiSHS18}
Zhuo Hui, Kalyan Sunkavalli, Sunil Hadap, and Aswin~C. Sankaranarayanan.
\newblock Illuminant spectra-based source separation using flash photography.
\newblock In {\em Proc.~CVPR}, 2018.

\bibitem{kingma2014adam}
Diederik~P Kingma and Jimmy Ba.
\newblock Adam: A method for stochastic optimization.
\newblock {\em arXiv preprint arXiv:1412.6980}, 2014.

\bibitem{kokkinos2019iterative}
Filippos Kokkinos and Stamatis Lefkimmiatis.
\newblock Iterative residual cnns for burst photography applications.
\newblock In {\em Proc.~CVPR}, 2019.

\bibitem{KrishnanF09}
Dilip Krishnan and Rob Fergus.
\newblock Dark flash photography.
\newblock {\em ACM Transactions on Graphics (TOG)}, 28(3):96, 2009.

\bibitem{LiL09}
Huibin Li and Feng Liu.
\newblock Image denoising via sparse and redundant representations over learned
  dictionaries in wavelet domain.
\newblock In {\em Proc.~International Conference on Image and Graphics (ICIG)},
  2009.

\bibitem{LiHA016}
Yijun Li, Jia{-}Bin Huang, Narendra Ahuja, and Ming{-}Hsuan Yang.
\newblock Deep joint image filtering.
\newblock In {\em Proc.~ECCV}, 2016.

\bibitem{LindenbaumFB94}
Michael Lindenbaum, M. Fischer, and Alfred~M. Bruckstein.
\newblock On gabor's contribution to image enhancement.
\newblock {\em Pattern Recognit.}, 27(1):1--8, 1994.

\bibitem{liu2018non}
Ding Liu, Bihan Wen, Yuchen Fan, Chen~Change Loy, and Thomas~S Huang.
\newblock Non-local recurrent network for image restoration.
\newblock In {\em Advances in Neural Information Processing Systems}, pages
  1680--1689, 2018.

\bibitem{liu2014fast}
Ziwei Liu, Lu Yuan, Xiaoou Tang, Matt Uyttendaele, and Jian Sun.
\newblock Fast burst images denoising.
\newblock {\em ACM Transactions on Graphics (TOG)}, 33(6):232, 2014.

\bibitem{mkpn}
Talmaj Marin{\v{c}}, Vignesh Srinivasan, Serhan G{\"u}l, Cornelius Hellge, and
  Wojciech Samek.
\newblock Multi-kernel prediction networks for denoising of burst images.
\newblock In {\em Proc.~ICIP}, 2019.

\bibitem{burstkpn}
Ben Mildenhall, Jonathan~T Barron, Jiawen Chen, Dillon Sharlet, Ren Ng, and
  Robert Carroll.
\newblock Burst denoising with kernel prediction networks.
\newblock In {\em Proc.~CVPR}, 2018.

\bibitem{PeronaM90}
Pietro Perona and Jitendra Malik.
\newblock Scale-space and edge detection using anisotropic diffusion.
\newblock {\em IEEE Transactions on pattern analysis and machine intelligence},
  12(7):629--639, 1990.

\bibitem{Petschnigg04}
Georg Petschnigg, Richard Szeliski, Maneesh Agrawala, Michael Cohen, Hugues
  Hoppe, and Kentaro Toyama.
\newblock Digital photography with flash and no-flash image pairs.
\newblock {\em ACM Transactions on Graphics (TOG)}, 23(3):664–672, Aug. 2004.

\bibitem{Rudin92}
Leonid~I Rudin, Stanley Osher, and Emad Fatemi.
\newblock Nonlinear total variation based noise removal algorithms.
\newblock {\em Physica D: nonlinear phenomena}, 60(1-4):259--268, 1992.

\bibitem{TomasiM98}
Carlo Tomasi and Roberto Manduchi.
\newblock Bilateral filtering for gray and color images.
\newblock In {\em Proc.~ICCV}, 1998.

\bibitem{WangXBC19}
Jian Wang, Tianfan Xue, Jonathan~T. Barron, and Jiawen Chen.
\newblock Stereoscopic dark flash for low-light photography.
\newblock In {\em Proc.~ICCP}, 2019.

\bibitem{wronski2019handheld}
Bartlomiej Wronski, Ignacio Garcia-Dorado, Manfred Ernst, Damien Kelly, Michael
  Krainin, Chia-Kai Liang, Marc Levoy, and Peyman Milanfar.
\newblock Handheld multi-frame super-resolution.
\newblock {\em ACM Transactions on Graphics (TOG)}, 38(4):1--18, 2019.

\bibitem{xia2018identifying}
Zhihao Xia and Ayan Chakrabarti.
\newblock Identifying recurring patterns with deep neural networks for natural
  image denoising.
\newblock In {\em Proc.~WACV}, 2020.

\bibitem{XiaPGSC20}
Zhihao Xia, Federico Perazzi, Micha{\"{e}}l Gharbi, Kalyan Sunkavalli, and Ayan
  Chakrabarti.
\newblock Basis prediction networks for effective burst denoising with large
  kernels.
\newblock In {\em Proc.~CVPR}, 2020.

\bibitem{xie2012denoising}
Junyuan Xie, Linli Xu, and Enhong Chen.
\newblock Image denoising and inpainting with deep neural networks.
\newblock In {\em Advances in Neural Information Processing Systems}, pages
  341--349, 2012.

\bibitem{yan2013crossfield}
Qiong Yan, Xiaoyong Shen, Li Xu, Shaojie Zhuo, Xiaopeng Zhang, Liang Shen, and
  Jiaya Jia.
\newblock Cross-field joint image restoration via scale map.
\newblock In {\em Proc.~ICCV}, 2013.

\bibitem{Yaroslavsky85}
Leonid~P Yaroslavsky.
\newblock Digital picture processing: an introduction.
\newblock {\em Applied Optics}, 25(18):3127, 1986.

\newpage

\bibitem{dncnn}
Kai Zhang, Wangmeng Zuo, Yunjin Chen, Deyu Meng, and Lei Zhang.
\newblock Beyond a gaussian denoiser: Residual learning of deep cnn for image
  denoising.
\newblock {\em IEEE Transactions on Image Processing}, 26(7):3142--3155, 2017.

\bibitem{ircnn}
Kai Zhang, Wangmeng Zuo, Shuhang Gu, and Lei Zhang.
\newblock Learning deep cnn denoiser prior for image restoration.
\newblock In {\em Proc.~CVPR}, 2017.

\bibitem{zhang2019residual}
Yulun Zhang, Kunpeng Li, Kai Li, Bineng Zhong, and Yun Fu.
\newblock Residual non-local attention networks for image restoration.
\newblock In {\em Proc.~ICLR}, 2018.

\bibitem{ZhuoGS10}
Shaojie Zhuo, Dong Guo, and Terence Sim.
\newblock Robust flash deblurring.
\newblock In {\em Proc.~CVPR}, 2010.

\end{thebibliography}
}

\clearpage
\pagenumbering{roman}

\onecolumn
\begin{center}
  ~\\\textbf{Supplementary Material}\\~
\end{center}

\appendix
\section{Architecture Details}

We describe our network architecture in detail in Table~\ref{tab:archi}. Our
architecture follows the encoder with dual decoder architecture
of~\cite{XiaPGSC20}, but changes the output of the global decoder to output a
basis with the two sets of kernels $\{(A_{j}, B_{j})\}$, each a
$K\times K\times 3$ kernel, and the per-pixel decoder outputs a scale map in
addition to the kernel coefficients.

\setcounter{table}{2}
\begin{table}[!h]
  \centering
  \begin{tabular}{llll}
    \toprule
    Name & Input & Layer & Output Size\\\midrule
    Input & - & - & H x W x 12\\\midrule
    \multicolumn{3}{l}{\em Encoder}\\\midrule
    Enc-0 & Input & 3x3 Conv & H x W x 64\\
    Enc-1-A & Enc-0 & 3x3 Conv & H x W x 64\\
    Enc-1-B & Enc-1-A & 3x3 Conv & H x W x 64\\
    Enc-1-C & Enc-1-B & 2x2 Stride 2 Max Pool & H/2 x W/2 x 64\\
    Enc-2-A & Enc-1-C & 3x3 Conv & H/2 x W/2 x 128\\
    Enc-2-B & Enc-2-A & 3x3 Conv & H/2 x W/2 x 128\\
    Enc-2-C & Enc-2-B & 2x2 Stride 2 Max Pool & H/4 x W/4 x 128\\
    Enc-3-A & Enc-2-C & 3x3 Conv & H/4 x W/4 x 256\\
    Enc-3-B & Enc-3-A & 3x3 Conv & H/4 x W/4 x 256\\
    Enc-3-C & Enc-3-B & 2x2 Stride 2 Max Pool & H/8 x W/8 x 256\\
    Enc-4-A & Enc-3-C & 3x3 Conv & H/8 x W/8 x 512\\
    Enc-4-B & Enc-4-A & 3x3 Conv & H/8 x W/8 x 512\\
    Enc-4-C & Enc-4-B & 2x2 Stride 2 Max Pool & H/16 x W/16 x 512\\
    Enc-5-A & Enc-4-C & 3x3 Conv & H/16 x W/16 x 1024\\
    Enc-5-B & Enc-5-A & 3x3 Conv & H/16 x W/16 x 1024\\
    Enc-5-C & Enc-5-B & 2x2 Stride 2 Max Pool & H/32 x W/32 x 1024\\
    Enc-F & Enc-5-C & 3x3 Conv & H/32 x W/32 x 1024\\
    Enc-Out & Enc-F & 3x3 Conv & H/32 x W/32 x 1024\\\midrule
    \multicolumn{3}{l}{\em Global Decoder}\\\midrule
    GDec-5-A & Bilinear-Up(GP(Enc-Out)) & 3x3 Conv & 2x2 x 512\\
    GDec-5-B & GDec-5-A, GP-R[2x2](Enc-5-B) & 3x3 Conv & 2 x 2 x 512\\
    GDec-5-C & GDec-5-B & 3x3 Conv & 2 x 2 x 512\\
    GDec-4-A & Bilinear-Up(GDec-5-C) & 3x3 Conv & 4 x 4 x 256\\
    GDec-4-B & GDec-4-A, GP-R[4x4](Enc-4-B) & 3x3 Conv & 4 x 4 x 256\\
    GDec-4-C & GDec-4-B & 3x3 Conv & 4 x 4 x 256\\
    GDec-3-A & Bilinear-Up(GDec-4C) & 3x3 Conv & 8 x 8 x 256\\
    GDec-3-B & GDec-3-A, GP-R[8,8](Enc-3-B) & 3x3 Conv & 8 x 8 x 256\\
    GDec-3-C & GDec-3-B & 3x3 Conv & 8 x 8 x 256\\
    GDec-2-A & Bilinear-Up(GDec-3-C) & 3x3 Conv & 16 x 16 x 128\\
    GDec-2-B & GDec-2-A, GP-R[16,16](Enc-2-B) & 3x3 Conv & 16 x 16 x 128\\
    GDec-2-C & GDec-2-B & 3x3 Conv & 16 x 16 x 128\\
    GDec-F-A & GDec-2-C & 2x2 Conv (Valid) & 15 x 15 x 128\\
    GDec-F-B & GDec-F-A & 3x3 Conv & 15 x 15 x 128\\
    Output: Basis & GDec-F-B & 3x3 Conv & 15 x 15 x (3*2*J)\\
    \toprule
  \end{tabular}\vspace{-0.9em}
  \caption{{\bf Our network architecture.} Bi-linear upsampling refers to
    upsampling the feature map by a factor of 2. GP refers to global average
    pooling, and GP-R[H', W'] to global average pooling followed by replicating
    spatially to size H' x W'. Multiple inputs are concatenated along the
    channel dimension before being passed to the convolution layer. All
    convolution layers use same padding unless otherwise
    specified.}\label{tab:archi}
\end{table}

\setcounter{table}{2}
\begin{table}[!h]
  \centering
  \begin{tabular}{llll}
    \toprule
    Name & Input & Layer & Output Size\\\midrule
    \multicolumn{3}{l}{\em Per-pixel Decoder}\\\midrule
    PDec-5-A & Bilinear-Up(Enc-Out) & 3x3 Conv & H/16 x W/16 x 512\\
    PDec-5-B & PDec-5-A, Enc-5-B & 3x3 Conv & H/16 x W/16 x 512\\
    PDec-5-C & PDec-5-B & 3x3 Conv & H/16 x W/16 x 512\\
    PDec-4-A & Bilinear-Up(PDec-5-C) & 3x3 Conv & H/8 x W/8 x 256\\
    PDec-4-B & PDec-4-A, Enc-4-B & 3x3 Conv & H/8 x W/8 x 256\\
    PDec-4-C & PDec-4-B & 3x3 Conv & H/8 x W/8 x 256\\
    PDec-3-A & Bilinear-Up(PDec-4-C) & 3x3 Conv & H/4 x W/4 x 128\\
    PDec-3-B & PDec-3-A, Enc-3-B & 3x3 Conv & H/4 x W/4 x 128\\
    PDec-3-C & PDec-3-B & 3x3 Conv & H/4 x W/4 x 128\\
    PDec-2-A & Bilinear-Up(PDec-3-C) & 3x3 Conv & H/2 x W/2 x 64\\
    PDec-2-B & PDec-2-A, Enc-2-B & 3x3 Conv & H/2 x W/2 x 64\\
    PDec-2-C & PDec-2-B & 3x3 Conv & H/2 x W/2 x 64\\
    PDec-1-A & Bilinear-Up(PDec-2-C) & 3x3 Conv & H x W x 64\\
    PDec-1-B & PDec-1-A, Enc-1-B & 3x3 Conv & H x W x 64\\
    PDec-1-C & PDec-1-B & 3x3 Conv & H x W x 64\\
    PDec-F-0 & PDec-1-C & 3x3 Conv & H x W x 64\\
    PDec-F-1 & PDec-F-0 & 3x3 Conv & H x W x 64\\
    Output: Coeffs + Scale map & PDec-F-1 & 3x3 Conv & H x W x (J + 3)\\\toprule
  \end{tabular}
  \caption{(continued) {\bf Our network architecture.}}
\end{table}

The output of the global decoder gives us $J=90$ pairs of kernels
$\{(A_{j},B_{j})\}$, and that of the per-pixel decoder both the coefficients
$C[n] \in \mathbb{R}^{J}$ and a scale map $G[n] \in \mathbb{R}^{3}$. As noted in
the paper, we set $J=90$ in our experiments.

\myparagraph{Baselines} Our baselines use similar architectures to our main
method to enable a fair comparison. For the no-flash single image input, we use
the above architecture to take only the no-flash image as input (6 channels: 3
for the image itself and 3 for noise deviation maps), and do not output a scale
map (i.e., our per-pixel decoder only outputs the $J$ channel coefficient map).
For the BPN~\cite{XiaPGSC20} entries in our table for the 2x burst of no-flash
images as well as for flash and no-flash denoising, we use the original
architecture from their paper. Specifically, our per-pixel decoder again does
not output a scale map, and the two sets of kernels are used to filter the two
input images which are then added---unlike our approach which combines the two
kernels to create a larger upsampled kernel that is only applied to the ambient,
followed by multiplication with the scale map. For the direct prediction and
KPN~\cite{burstkpn}, we use this architecture without the global decoder. For
direct prediction, the per-pixel decoder just outputs a 3-channel map that is
treated as a residual and added to the noisy no-flash input to yield the final
denoised output. For KPN, the per-pixel decoder outputs a 150-channel output:
these are interpreted as two $5\times 5$ kernels per color channel, to be
applied to the flash and no-flash pairs.

\section{Additional Results}

\myparagraph{Qualitative Results} We show comparison results on more images in Figure~\ref{fig:more}.
\setcounter{figure}{6}
\begin{figure*}[!t]
  \centering
  \includegraphics[width=0.84\textwidth]{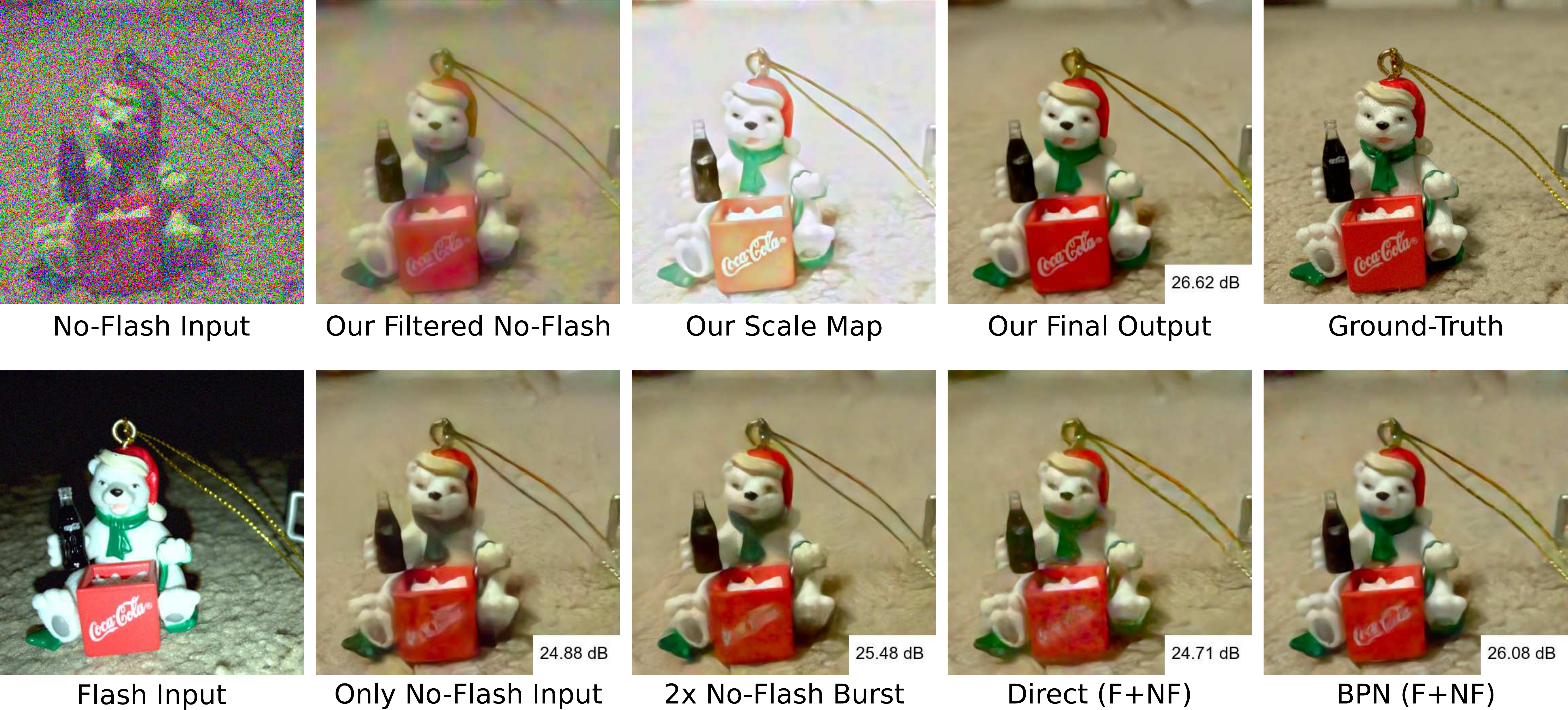}\\~\\
  \includegraphics[width=0.84\textwidth]{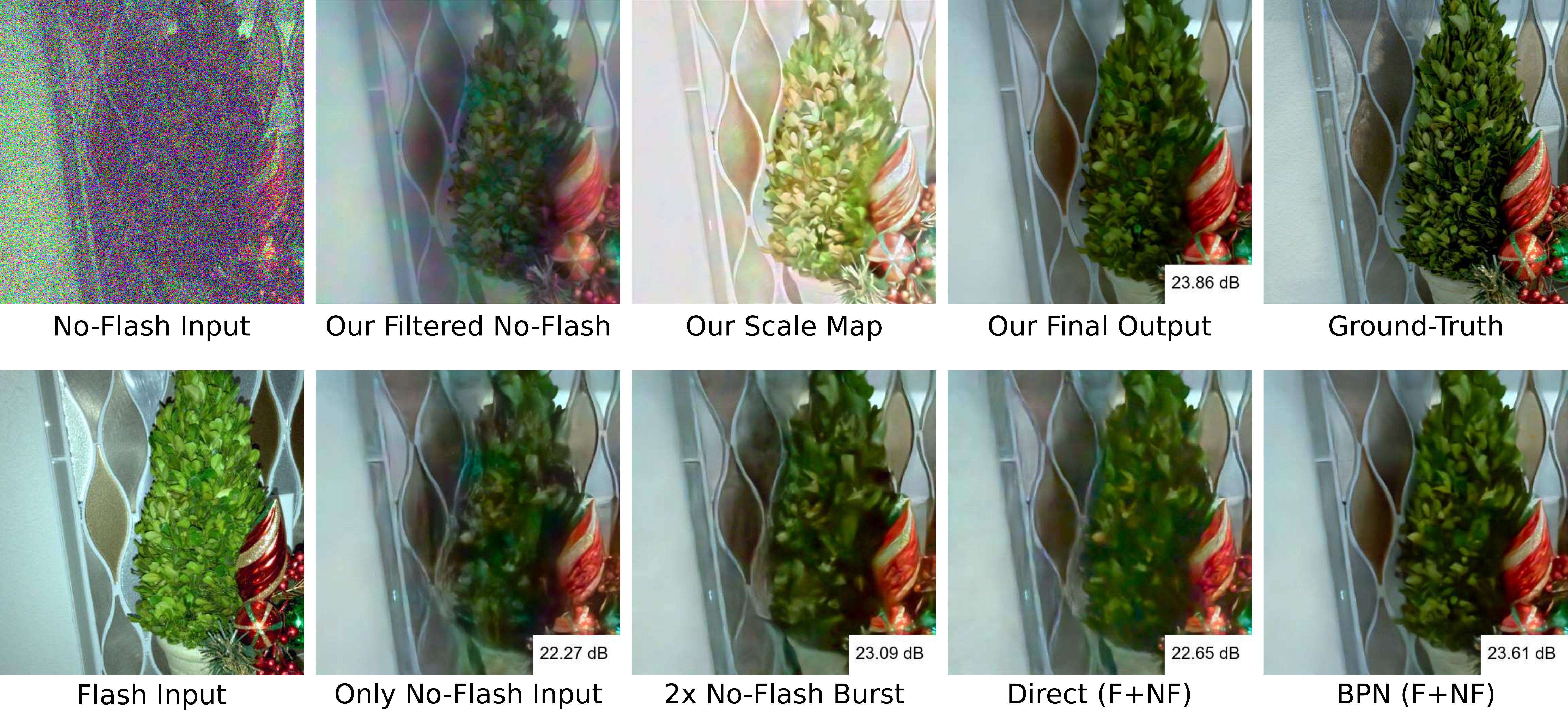}\\~\\
  \includegraphics[width=0.84\textwidth]{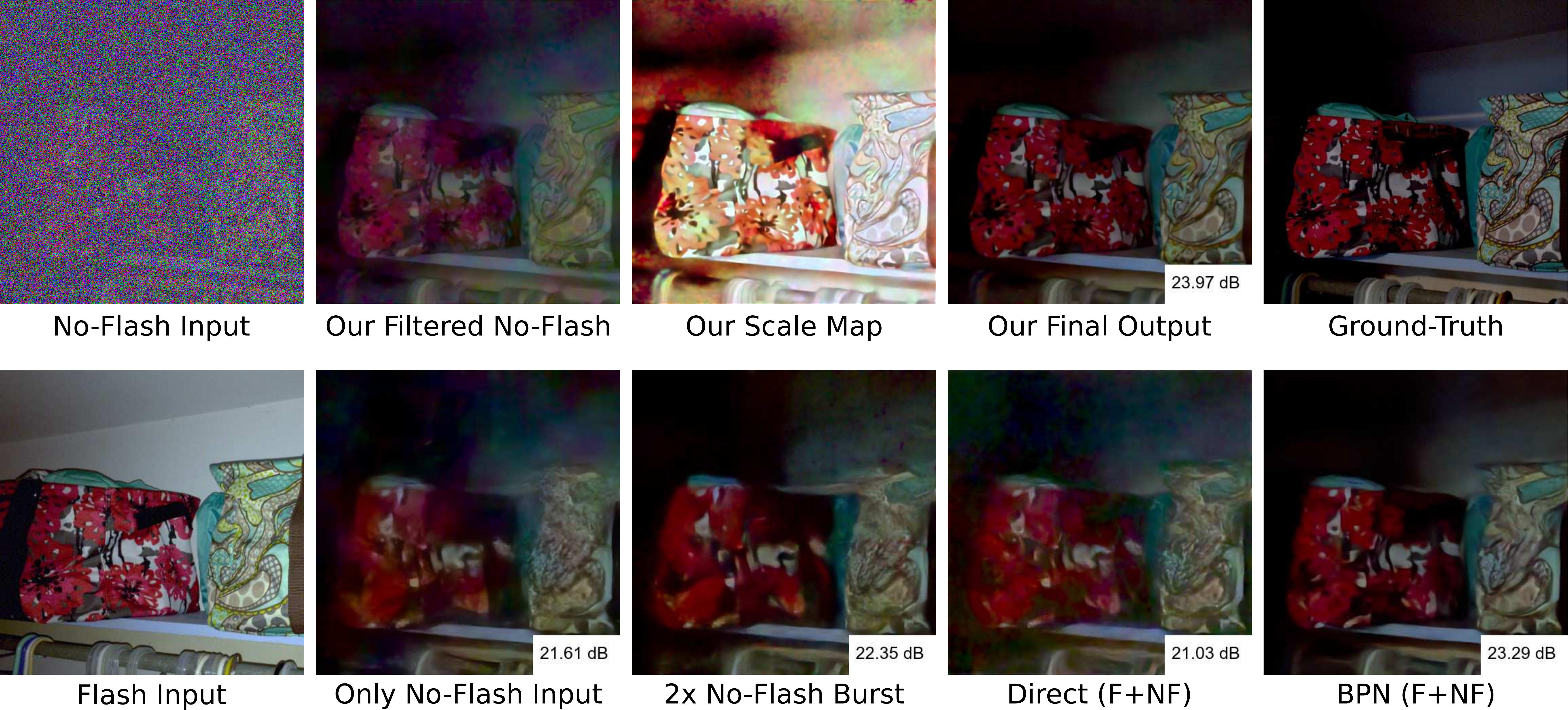}
  \caption{\label{fig:more} {\bf More qualitative comparisons.}}
\end{figure*}
\setcounter{figure}{6}
\begin{figure*}[!t]
  \centering
  \includegraphics[width=0.84\textwidth]{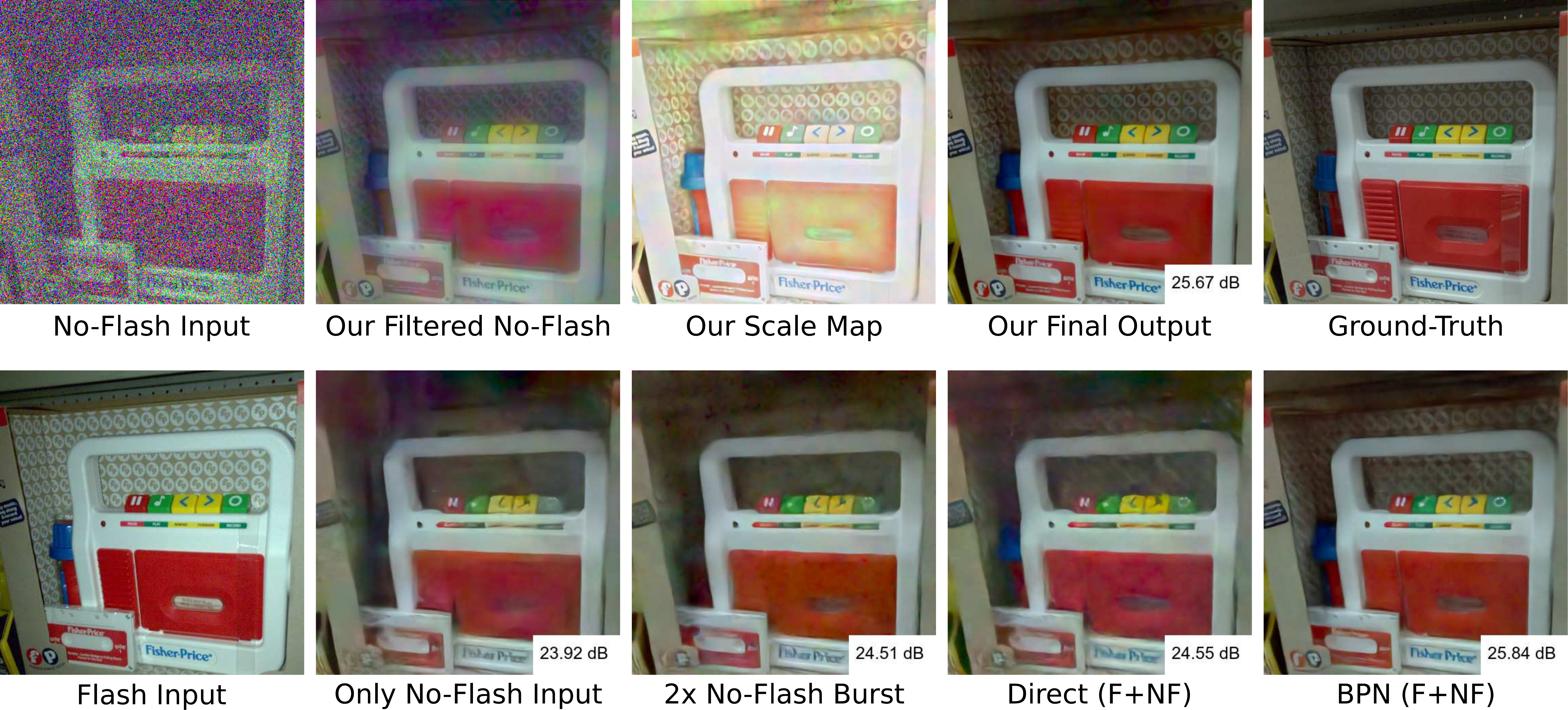}\\~\\
  \includegraphics[width=0.84\textwidth]{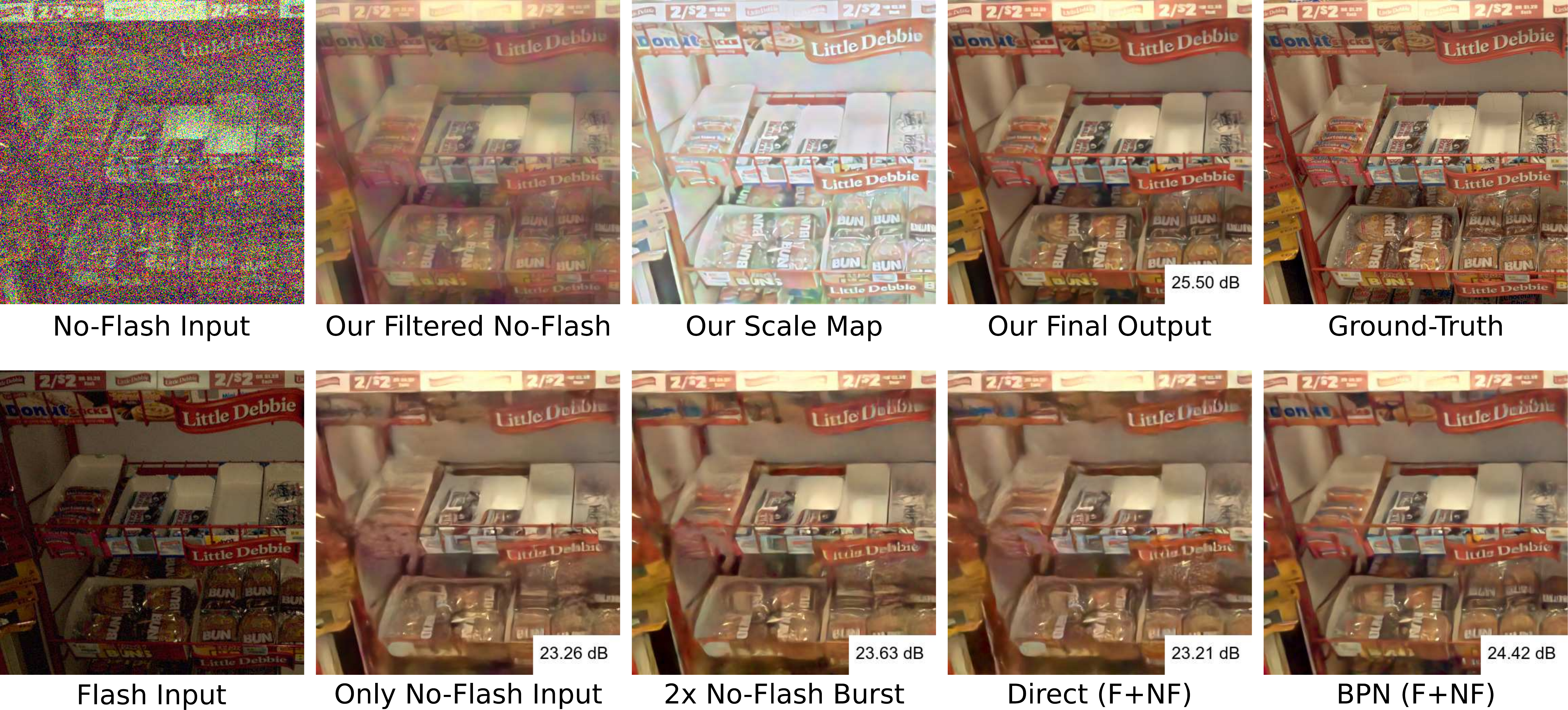}\\~\\
  \includegraphics[width=0.84\textwidth]{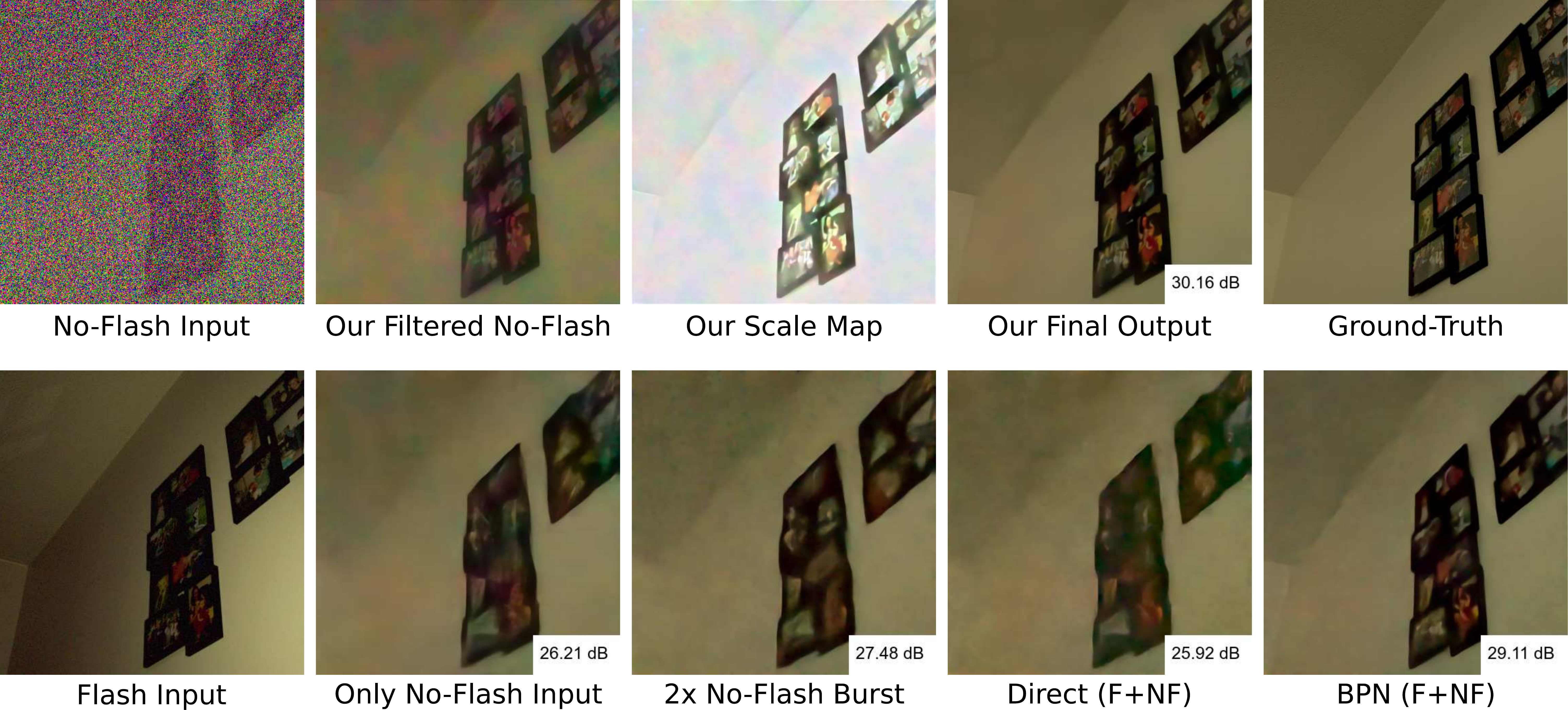}
  \caption{(continued) {\bf More qualitative comparisons.} }
\end{figure*}
\setcounter{figure}{6}
\begin{figure*}[!t]
  \centering
  \includegraphics[width=0.84\textwidth]{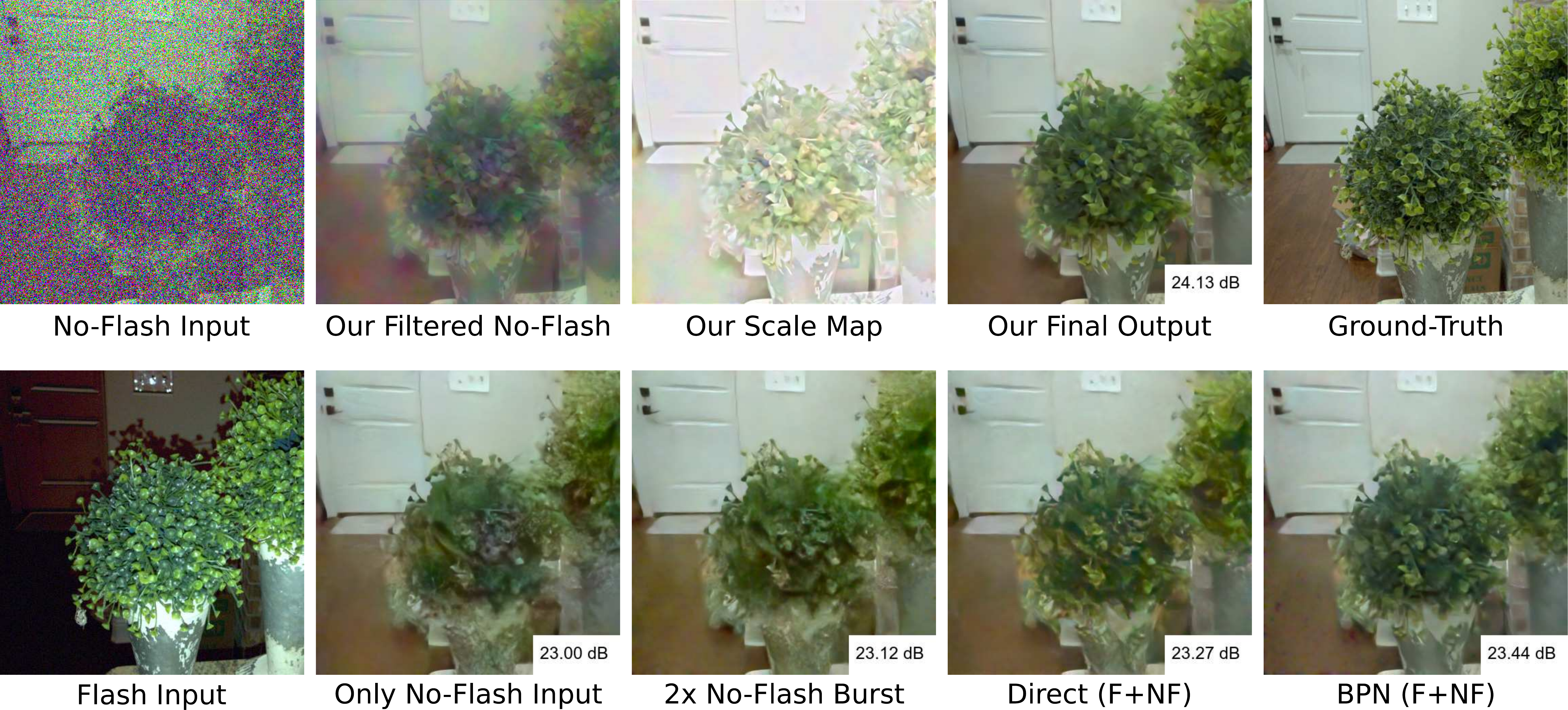}\\~\\
  \includegraphics[width=0.84\textwidth]{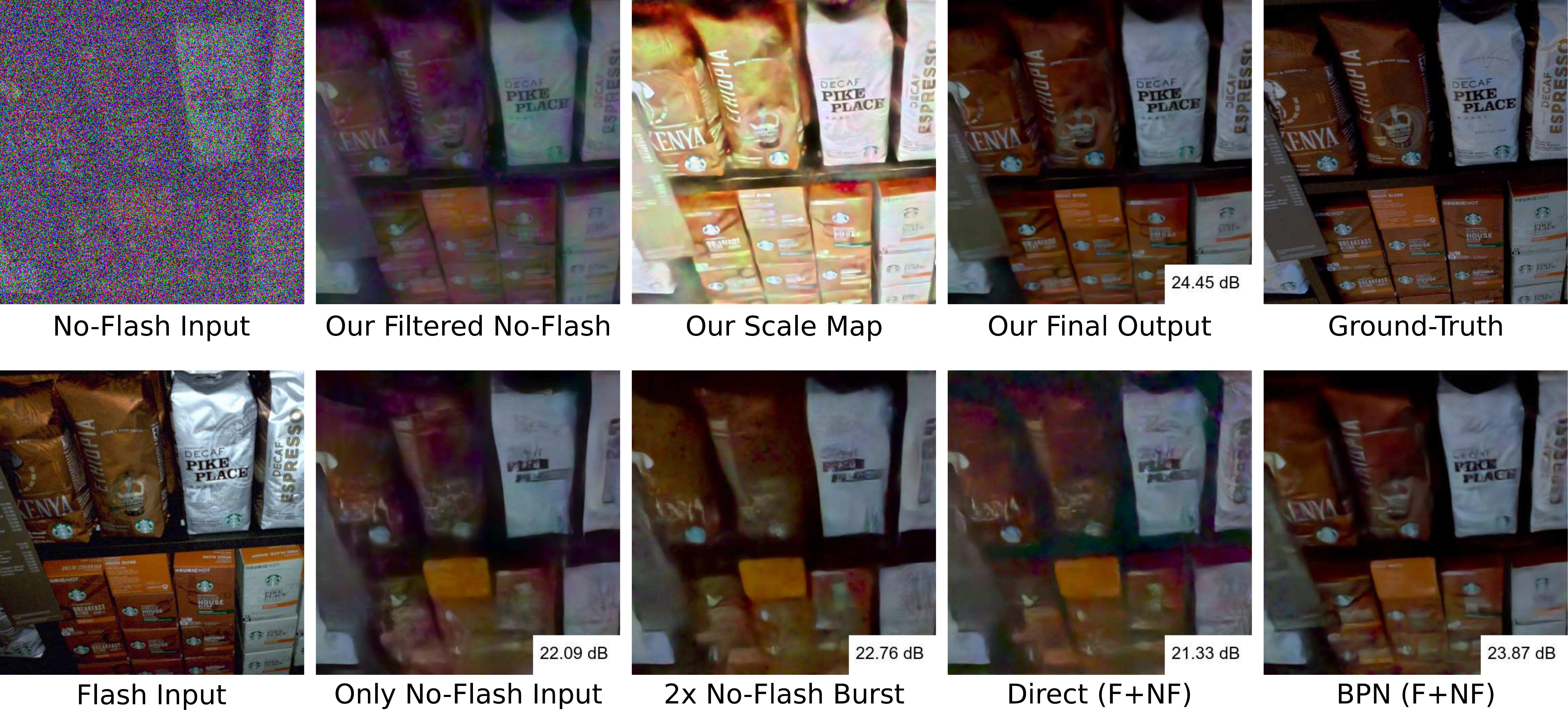}\\~\\
  \includegraphics[width=0.84\textwidth]{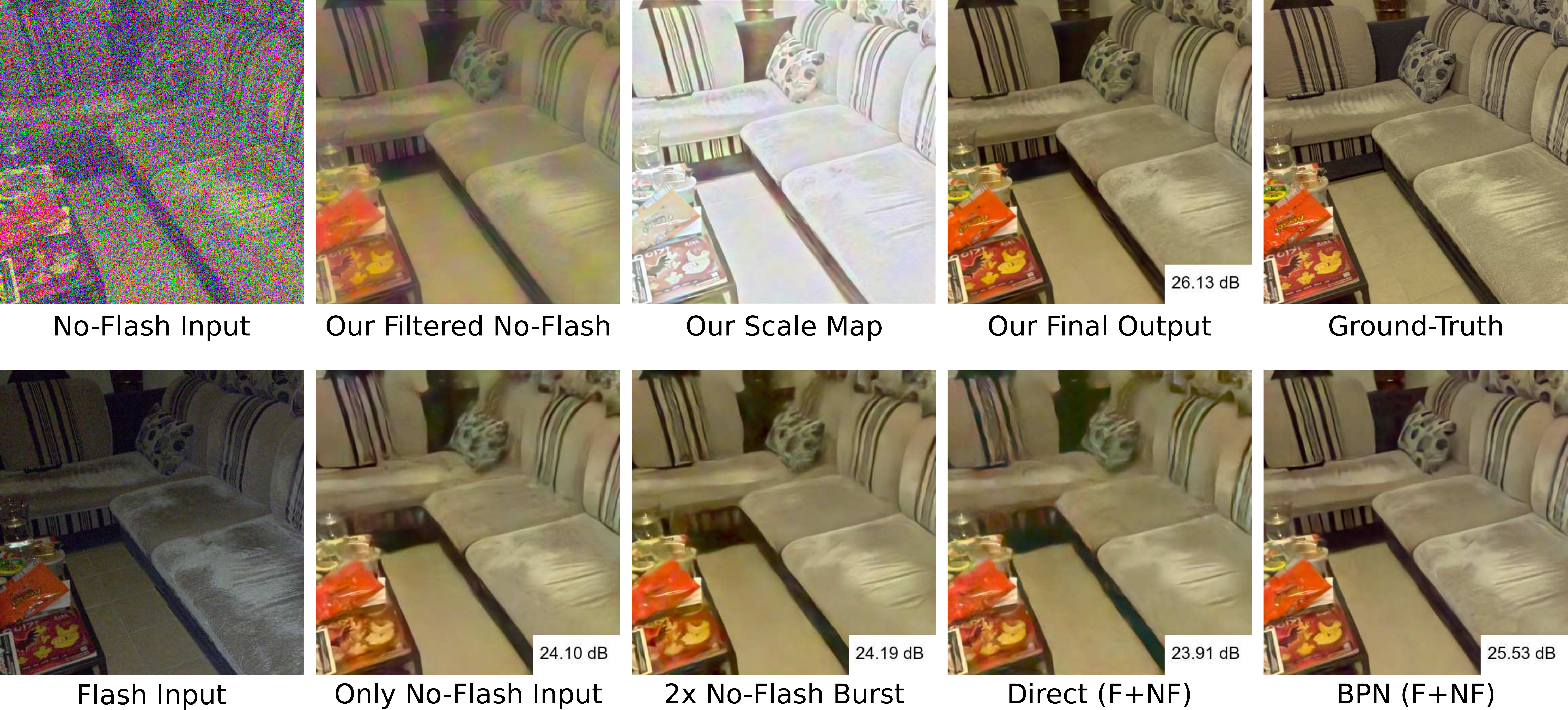}
  \caption{(continued) {\bf More qualitative comparisons.} }
\end{figure*}

\begin{figure*}[!t]
  \centering
  \includegraphics[width=\textwidth]{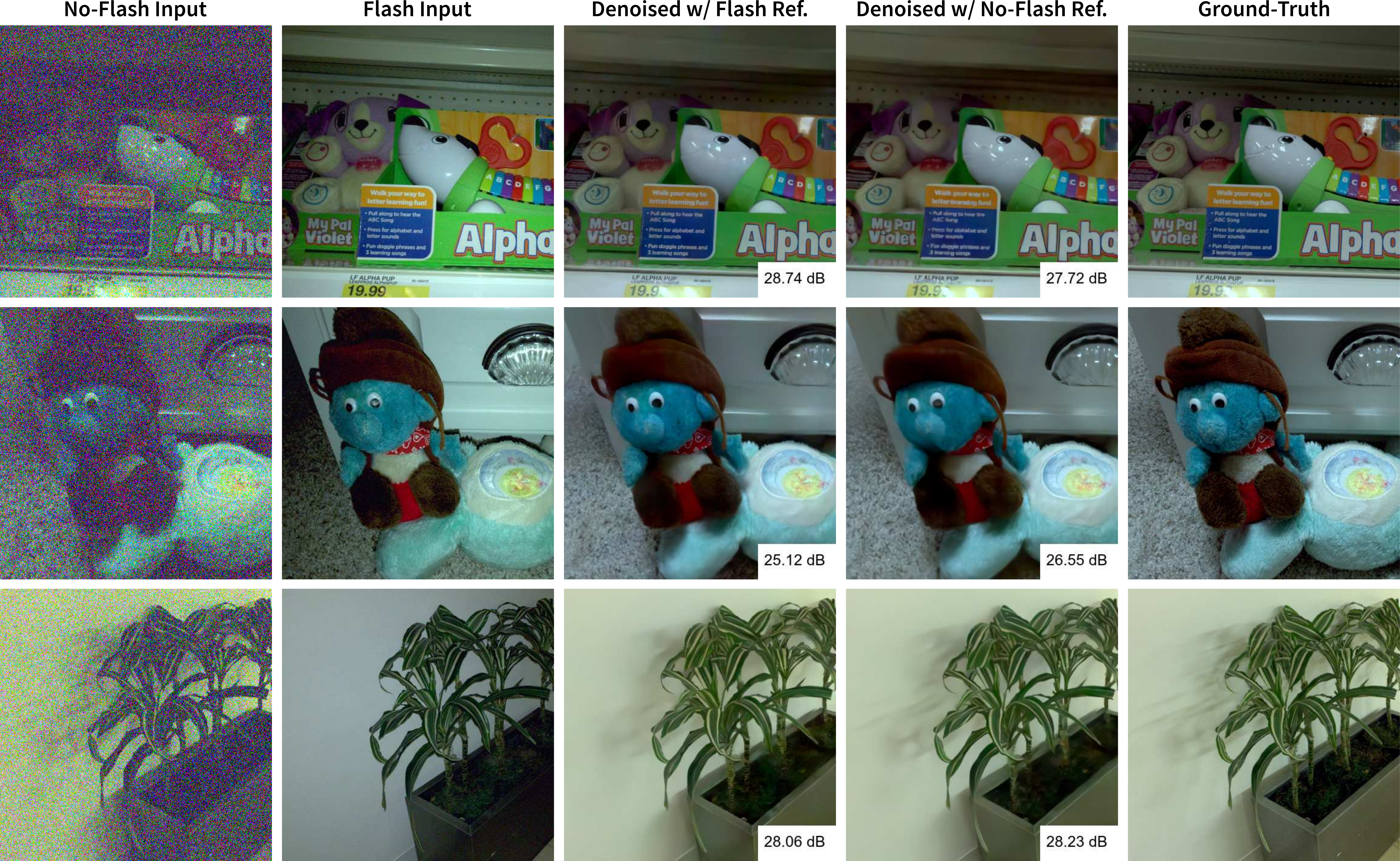}
  \caption{\label{fig:ftarget} {\bf Flash vs No-Flash as reference.} We show
    qualitative comparisons of results (at 50x dimmed light level) for our
    method using the flash input vs.\ the no-flash input as geometric reference.
    While using the no-flash input as reference does better on average, in some
    examples, using the flash as reference can lead to better reconstruction of
    high frequency detail (first row). In other cases, even though the
    flash-reference results are quantitatively, the difference is due to errors
    in reconstructing shading which are often less perceptually obvious. This is
    the case in the bottom two rows, although in the last row, we can see that
    the flash reference output has a blurrier reconstruction of the shadows on
    the wall. }
\end{figure*}

\myparagraph{No-Flash vs.\ Flash as reference} As noted in the paper, using the
no-flash image as the geometric reference leads to better performance at all but
the darkest light level. This is true not just for our approach, but also the
other baselines we considered for denoising with flash and no-flash image pairs
as input. We report the performance of these methods when using flash as
reference in Table~\ref{tab:ftarget}, and find that like in the case of no-flash
reference, our approach yields superior reconstructions. Although slightly worse
on average, we find that using flash as reference can sometimes lead to superior
reconstruction of high-frequency details for some images compared to the
no-flash reference, and even when the results are quantitatively worse, this is
because of misalignment of low-frequency shading that is often not perceptible.
We illustrate this with qualitative comparisons in Fig.~\ref{fig:ftarget}.

\begin{table*}[!t]
  \centering
  \begin{tabular}{cl c cc c cc c cc c cc}
    \toprule
    &\multirow{2}{*}{\textbf{Method}} && \multicolumn{2}{c}{\bf 100x Dimmed} && \multicolumn{2}{c}{\bf 50x Dimmed} && \multicolumn{2}{c}{\bf 25x Dimmed} && \multicolumn{2}{c}{\bf 12.5x Dimmed}\\
    &&& PSNR & SSIM && PSNR & SSIM && PSNR & SSIM && PSNR & SSIM\\\midrule
    &Direct Prediction    && 24.15 dB & 0.779 && 26.00 dB & 0.810 && 27.31 dB & 0.836 && 28.17 dB & 0.856\\
    &KPN~\cite{burstkpn}  && 25.51 dB & 0.820 && 27.43 dB & 0.852 && 28.78 dB & 0.874 && 29.74 dB & 0.889\\
    &BPN~\cite{XiaPGSC20} && 26.23 dB & 0.831 && 27.83 dB & 0.857 && 29.08 dB & 0.877 && 30.00 dB & 0.891\\
    &\textbf{Ours}        && \bf 26.83 dB & \bf 0.843 && \bf 28.39 dB & \bf 0.866 && \bf 29.55 dB & \bf 0.883 && \bf 30.45 dB & \bf 0.897\\\bottomrule

  \end{tabular}
  \caption{\label{tab:ftarget}\bf{Quantitative Results using flash image as
      geometric reference for all methods.}}
\end{table*}

\myparagraph{Other Noise Levels} In addition to the original values of read and
shot noise variances used in the main results table, Table~\ref{tab:morenoise}
reports the performance of our method, and other baselines for denoising flash
and no-flash pairs, for three additional sets of noise parameters at one of the
light levels.

\begin{table*}[!t]
  \centering
  \begin{tabular}{cl c cc c cc c cc}
    \toprule
    &\multirow{2}{*}{\textbf{Noise Parameters}} && \multicolumn{2}{c}{$\log([\sigma_{r}, \sigma_{s}])$ = [-2.8, -4.0]} && \multicolumn{2}{c}{$\log([\sigma_{r}, \sigma_{s}])$ = [-2.4, -3.2]} && \multicolumn{2}{c}{$\log([\sigma_{r}, \sigma_{s}])$ = [-2.2, -2.8]}\\
    &&& PSNR & SSIM && PSNR & SSIM && PSNR & SSIM\\\midrule
    &Direct Prediction    && 28.44 dB & 0.845 && 25.54 dB & 0.787 && 23.88 dB & 0.754\\
    &KPN~\cite{burstkpn}  && 29.01 dB & 0.870 && 26.82 dB & 0.832 && 25.59 dB & 0.808\\
    &BPN~\cite{XiaPGSC20} && 29.18 dB & 0.870 && 26.86 dB & 0.829 && 25.61 dB & 0.805\\
    &\textbf{Ours}        && \bf 29.65 dB & \bf 0.876 && \bf 27.47 dB & \bf 0.842 && \bf 26.26 dB & \bf 0.821\\\bottomrule

  \end{tabular}
  \caption{\label{tab:morenoise} {\bf Noise levels.} Performance of different
    approaches to denoising flash and no-flash pairs, at 50x dimmed light
    levels, with three additional noise levels (increasing from left to right).
    Note that the noise level used in the main results in the paper was between
    the first and second level above.}
\end{table*}

\myparagraph{Burst Denoising} We compared to using a burst of two no-flash
images in the paper, as a means of evaluating the relative benefit of a the
second image being taken with vs.\ without a flash. In both cases the second
image provides additional information---a second no-flash image has high noise
(though a different realization of noise than the first image), while a flash
image has a much higher signal-to-noise ratio but entirely different shading.
Our results showed that in this context, a second image is beneficial.

But more generally, burst photography (which typically involves a larger number
of images) has its own relative strengths and weaknesses when compared to using
flash and no-flash pairs, as a means of imaging in low-light. Burst denoising
with longer bursts may well be a preferable option in the presence of moderate
motion, or when using a flash is not an option (for example, when most objects
in the scene are far away and can not be illuminated with a flash). Conversely,
the use of a flash and no-flash pair is preferable in much lower light, in
scenes where most of the scene \emph{can} be well-illuminated with a flash, when
camera or scene motion may cause significant misalignment across a large
sequence of images, or when memory or computational constraints prevent
capturing a larger sequence of images.

\begin{table*}[!t]
  \centering
  \begin{tabular}{l  cc c cc c cc c cc}
    \toprule
    \multirow{2}{*}{\textbf{Method}} & \multicolumn{2}{c}{\bf 100x Dimmed} && \multicolumn{2}{c}{\bf 50x Dimmed} && \multicolumn{2}{c}{\bf 25x Dimmed} && \multicolumn{2}{c}{\bf 12.5x Dimmed}\\
    & PSNR & SSIM && PSNR & SSIM && PSNR & SSIM && PSNR & SSIM\\\midrule
    2x No-flash (BPN~\cite{XiaPGSC20})    & 25.58 dB & 0.796 && 27.75 dB & 0.839 && 29.65 dB & 0.874 && 31.21 dB & 0.899\\
    Flash and No-flash (Ours)             & 26.75 dB & 0.829 && 28.56 dB & 0.860 && 30.14 dB & 0.884 && 31.52 dB & 0.903\\
    5x No-flash (BPN~\cite{XiaPGSC20})    & 25.90 dB & 0.788 && 27.84 dB & 0.832 && 29.54 dB & 0.867 && 31.03 dB & 0.894\\
    Flash and 4x No-flash (Ours)          & 26.82 dB & 0.822 && 28.60 dB & 0.856 && 30.21 dB & 0.883 && 31.60 dB & 0.904\\
    7x No-flash (BPN~\cite{XiaPGSC20})    & 26.00 dB & 0.792 && 27.89 dB & 0.834 && 29.57 dB & 0.868 && 31.05 dB & 0.894\\
    Flash and 6x No-flash (Ours)          & 26.80 dB & 0.820 && 28.60 dB & 0.855 && 30.20 dB & 0.882 && 31.59 dB & 0.903\\\bottomrule
  \end{tabular}
  \caption{\textbf{\label{tab:lburst} Performance with larger bursts.} We
    compare the original results of 2x no-flash burst with BPN~\cite{XiaPGSC20}
    and flash and no-flash denoising with our method, to denoising larger bursts
    of no-flash images of length 5 and 7 with BPN, as well as using a modified
    version of our method with bursts of the same length where one of the images
    is captured with a flash.}
\end{table*}

While the question of what acquisition strategy to use will depend on the
environment and platform and is beyond the scope of this paper, we present a
comparison in Table~\ref{tab:lburst} to provide further intuition to the reader.
We compare burst denoising, using BPN~\cite{XiaPGSC20}, with larger bursts of 5
and 7 images on our dataset---we use the same noise and dimming models to
generate a larger burst of no-flash images. These sequences are mis-aligned
using our randomly sampled homographies, with the homographies applied
sequentially---thus the first and last image of a sequence will have a larger
misalignment on average than two subsequent pairs, and so we use the image in
the middle of the sequence as reference. We compare these results to using our
method when denoising bursts of the same size, where one (the last) image is
taken with a flash and the rest without (again using the middle no-flash image
as reference). Here, our network predicts kernels to be used to filter and sum
all the no-flash images, which is then multiplied with our scale map. Because
memory constraints, we do not use kernel upsampling in these experiments, and
predict only a basis of $15\times 15$ kernels (one for each channel of each
no-flash image).

Our results show that in the light and motion settings we consider, larger
no-flash bursts only have a modest improvement over a pair of two no-flash
images at lower light levels, although they perform slightly worse comparatively
at higher light levels (this is likely because the networks are trained over a
range of light levels, and tend to oversmooth to handle the lowest end of that
range). Our method, when using a burst of the same size with one as a flash
image, performs better than pure no-flash bursts, but also with only modest
improvements over a flash and no-flash pair (note that in this case, the
misalignment between the flash image and reference frame is greater than for a
flash and no-flash pair that are taken in sequence). These results suggest that
when using burst photography in settings where it is advantageous, it may be
worth capturing one image of that burst with a flash.

\end{document}